\documentclass[a4paper,twoside]{article}

\usepackage{epsfig}
\usepackage{subcaption}
\usepackage{calc}
\usepackage{amssymb}
\usepackage{amstext}
\usepackage{amsmath}
\usepackage{amsthm}
\usepackage{multicol}
\usepackage{pslatex}
\usepackage{apalike}
\usepackage[bottom]{footmisc}

\usepackage[utf8]{inputenc}  
\usepackage[T1]{fontenc} 
\usepackage[colorlinks=true,citecolor=blue]{hyperref}  
\usepackage{booktabs}  
\usepackage{amsfonts,amsmath,amssymb,bm,bbm}
\usepackage{algorithm}
\usepackage{nicefrac}
\usepackage{microtype}  
\usepackage{xcolor}
\usepackage{graphicx}
\usepackage[mathscr]{euscript}
\usepackage{enumitem}
\usepackage{pdflscape}
\usepackage{multirow}

\usepackage[noend]{algpseudocode}

\newcommand{\x}{\mathbf{x}}
\newcommand{\y}{\mathbf{y}}
\newcommand{\RR}{\mathbb{R}}

\newcommand{\EE}{\mathbb{E}}
\newcommand{\1}{\mathbbm{1}}
\newcommand{\wh}[1]{\widehat{#1}}

\newcommand{\der}{\partial}
\newcommand{\rme}{\mathrm{e}}
\newcommand{\rmd}{\mathrm{d}}
\def\ba#1\ea{\begin{align}#1\end{align}}
\def\mkakko#1{\left(#1\right)}
\def\ckakko#1{\left\{#1\right\}}
\def\kkakko#1{\left[#1\right]}

\newcommand{\Xb}{\mathscr{X}_{\mathrm{b}}}
\newcommand{\Pb}{P_{\mathrm{b}}}
\newcommand{\Nb}{N_{\mathrm{b}}}
\renewcommand{\a}{\bm{a}}
\renewcommand{\b}{\mathbf{b}}
\renewcommand{\c}{\mathbf{c}}
\newcommand{\p}{\mathbf{p}}
\newcommand{\q}{\mathbf{q}}

\renewcommand{\L}{\mathscr{L}}
\renewcommand{\epsilon}{\varepsilon}
\newcommand{\ALG}{\text{DN+}}

\def\old#1{}
\def\new#1{{#1}}

\usepackage{SCITEPRESS}

\begin{document}

\title{Sample-based Uncertainty Quantification with a Single 
Deterministic Neural Network}

\author{\authorname{Takuya Kanazawa and Chetan Gupta}
\affiliation{Industrial AI Lab, Hitachi America, Ltd.~R\&D, Santa Clara, CA 95054}
\email{\{takuya.kanazawa, chetan.gupta\}@hal.hitachi.com}
}


\abstract{Development of an accurate, flexible, and numerically efficient uncertainty quantification (UQ) method is one of fundamental challenges in machine learning. Previously, a UQ method called DISCO Nets has been proposed (Bouchacourt et al., 2016), which trains a neural network by minimizing the energy score. In this method, a random noise vector in $\RR^{10\text{--}100}$ is concatenated with the original input vector in order to produce a diverse ensemble forecast despite using a single neural network. While this method has shown promising performance on a hand pose estimation task in computer vision, it remained unexplored whether this method works as nicely for regression on tabular data, and how it competes with more recent advanced UQ methods such as NGBoost. In this paper, we propose an improved neural architecture of DISCO Nets that admits faster and more stable training while only using a compact noise vector of dimension $\sim \mathcal{O}(1)$. We benchmark this approach on miscellaneous real-world tabular datasets and confirm that it is competitive with or even superior to standard UQ baselines.  Moreover we observe that it exhibits better point forecast performance than a neural network of the same size trained with the conventional mean squared error. As another advantage of the proposed method, we show that local feature importance computation methods such as SHAP can be easily applied to any subregion of the predictive distribution. A new elementary proof for the validity of using the energy score to learn predictive distributions is also provided.}

\onecolumn \maketitle \normalsize \setcounter{footnote}{0} \vfill

\section{\uppercase{Introduction}}

In real-world applications of artificial intelligence (AI) and machine learning (ML), it is becoming essential to estimate uncertainty of predictions made by AI/ML models. This is especially true in high-stakes areas such as health care and autonomous driving, where inadvertent decisions can cause fatal damages. While traditional methods for uncertainty quantification (UQ) such as bootstrapping and quantile regression can be partly applied to modern AI/ML models, the rapid progress especially in the field of deep neural networks calls for a development of novel UQ methodologies \cite{ABDAR2021243,Gawlikowski2021}. 

In this paper, we revisit a UQ method for neural networks (NN) on regression tasks. This method, called DISCO Nets (DISsimilarity COefficient Networks) \cite{disco2016}, enables us to estimate uncertainty of a prediction in a fully nonparametric manner by using just a single deterministic NN (see also \cite{Harakeh2021,Pacchiardi2021} for related studies). Unlike Bayesian NN and Gaussian processes, DISCO Net does not encounter computational bottlenecks when it is scaled to a large problem. In addition, it can model a posterior distribution in more than one dimension straightforwardly, in contrast to conventional quantile regression-based methods that do not trivially generalize to higher dimensions. 

Despite its flexibility and versatility, however, DISCO Net has not gained popularity comparable to other UQ methods such as Monte Carlo dropout \cite{Gal2016}. There could be multiple reasons for that. First, DISCO Net belongs to a class of NN called implicit generative networks, which are generally difficult to train \cite{Tagasovska2019}. Second, understanding the theoretical underpinning of DISCO Net requires sophisticated mathematics and statistics of scoring rules of distributions, which makes the method unfamiliar and less approachable for data science practitioners in industry. Third, while it is widely known that there are two types of uncertainty in ML called \emph{aleatoric uncertainty} and \emph{epistemic uncertainty} \cite{ABDAR2021243,Gawlikowski2021,HW2021}, it is not totally obvious which of these uncertainties is estimated by DISCO Net. Fourth, DISCO Net has so far been primarily benchmarked on a limited range of tasks in computer vision, and evidence of its favorable performance on learning tasks on tabular data has been missing in the literature, despite abundance of tabular data in industry. 

We wish to address all these points in this paper. First, we propose an improved NN architecture of DISCO Net. The original DISCO Net assumed that a noise vector was simply concatenated to the input vector before being fed to the NN. However, empirically, the training of NN with this method is found to be difficult. The original work \cite{disco2016} has overcome this difficulty by using a very high ($\sim 200$) dimensional noise vector, but it consequently requires a quite large NN and thus increases the training time. We rather propose to simply insert an embedding layer of a noise vector. This trick leads to fast and stable learning \emph{even for a scalar noise} and boosts computational efficiency substantially. Second, we provide a rudimentary analytical proof that training of NN using the energy score allows to learn the correct predictive distribution, at least when the batch size in stochastic gradient descent training is large enough. We hope this will make DISCO Net more accessible and trustworthy for practical data scientists.  Third, we numerically verify that DISCO Net misses epistemic uncertainty, although it is capable of capturing aleatoric uncertainty quite well. We propose to use an oscillatory activation function in DISCO Net to capture epistemic uncertainty. Fourth, we conduct extensive numerical experiments to test the effectiveness of the proposed approach for tabular datasets. We use 10 real-world datasets and show that the UQ capability of the present approach is competitive with or even superior to popular baseline methods.

\begin{figure}[t]
	\includegraphics[height=.197\textwidth]{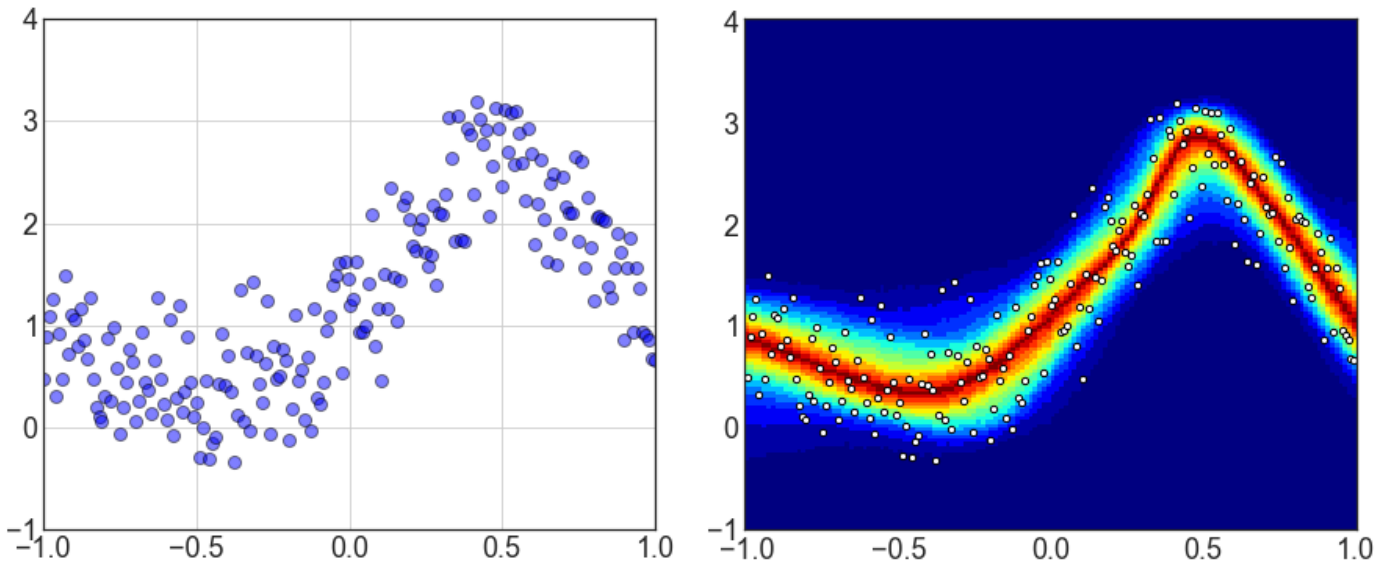}
	\caption{\label{fg:24fds}Left: random data points. Right: the result of applying {\ALG} to the same data. Aleatoric uncertainty is quantified appropriately. See section~\ref{sc:hgf98dfbv} for more detail.}
\end{figure}

The enhanced DISCO Net approach proposed in this work will be referred to as {\ALG} in the remainder of this paper. 

In section~\ref{sc:dsdf4} we summarize preceding works on UQ in machine learning. In section~\ref{sc:09e224987} we provide the background of this research, introducing concepts such as the energy score and CRPS. In section~\ref{sc:89dsuf} we review DISCO Nets. Section~\ref{sc:hgf98dfbv} is about our main contributions, i.e., improvements on DISCO Nets, discussions on numerical experiments, and comparison with baselines. We conclude in section~\ref{sc:cdosc7897}. A theoretical discussion on the validity of UQ training using the energy score is relegated to the appendix, owing to its technical nature.

\section{\uppercase{Related work}\label{sc:dsdf4}}

In a classification problem, the uncertainty of an ML model may be concisely represented by class probabilities that sum to unity. In contrast, the uncertainty in regression is more complicated. Many existing UQ methods for regression calculate only the $X$\% confidence interval, where $X\in\{95,99\}$ are among the common choices. Although such a single interval estimate is quite useful from a practical point of view, it lacks detailed information on the shape of the posterior distribution $p(\y|\x)$, where $\x$ denotes the input variable and $\y$ the output variable. Classical approaches such as the Gaussian process regression \cite{RWbook} can represent the posterior as a multivariate Gaussian distribution, but fails when the noise is heteroscedastic and is unable to express multimodal posterior distributions. 

A variety of improved UQ methods for deep NN have been proposed in the literature \cite{ABDAR2021243,Gawlikowski2021}.%
\footnote{Here we will focus on work other than DISCO Net \cite{disco2016,Harakeh2021,Pacchiardi2021} because the latter was discussed in great detail in Introduction.} Pivotal examples include Bayesian NN \cite{Lampinen2001}, quantile regression NN \cite{Cannon2011}, NN that model Gaussian mixtures \cite{BishopMDN}, ensembles of NN \cite{DE,Pearce2020}, Monte Carlo dropout \cite{Gal2016}, and direct UQ approaches \cite{DEUP}. These methods effectively work to quantify either aleatoric uncertainty, epistemic uncertainty, or both \cite{HW2021}. Aleatoric uncertainty represents inherent stochasticity of the response variable, whereas epistemic uncertainty comes from limitation of knowledge and can be decreased by gathering more data.

Recently, normalizing flows (NF) \cite{9089305,Papamakarios2021} have emerged as a versatile and flexible tool for UQ. NF is a generative model that uses a composition of multiple differentiable bijective maps modeled by NN to transform a simple (such as uniform or Gaussian) distribution to a more complex distribution of real data. Examples of probabilistic forecast based on NF can be found in \cite{Sick2020,Charpentier2020,Dumas2021,Sendera2021,Jamgochian2022,Rittler2022,Marz2022,Arpogaus2022,Cramer2022}. 

Ensembles from a single NN (known as \emph{implicit NN ensembles}) have been studied in \cite{Huang2016,Huang2017,Tagasovska2019,Maddox2019,Antoran2020}. While \cite{Huang2016,Antoran2020} propose a NN with a probabilistic depth, \cite{Huang2017,Maddox2019} suggest to use information in a stochastic gradient descent trajectory of a single NN to construct ensembles. 

Gradient boosting decision trees are among the most popular ML models, which often outperform NN as a point forecaster on benchmark tests with tabular data. Recently, probabilistic forecasts in gradient boosting decision trees have been studied in \cite{Duan2020,Sprangers2021,Brophy2022}. While \cite{Duan2020,Sprangers2021} require fitting a parametric distribution (such as Gaussian or Weibull) to the data, \cite{Brophy2022} allows to produce a more flexible, nonparametric distributional forecast. 

\section{\uppercase{Background}\label{sc:09e224987}}

\subsection{Distance between Probability Distributions}

The discrepancy (distance) between two continuous probability distributions can be quantified using a variety of metrics such as $f$-divergences. One of the popular metrics in machine learning is the \emph{maximum mean discrepancy} (MMD) \cite{Gretton2012}
\ba
	\hspace{-3mm}
	\mathrm{MMD}[\mathscr{F},p,q]
	& := \underset{f\in\mathscr{F}}{\mathrm{sup}}
	\mkakko{\EE_{x\sim p}[f(x)]-\EE_{y\sim q}[f(y)]}
\ea
where $p$ and $q$ are distributions and $\mathscr{F}$ is a class of real-valued functions $f:\mathscr{X}\to \RR$. When $\mathscr{F}$ is a reproducing kernel Hilbert space, there exists a kernel function $k:\mathscr{X}\times\mathscr{X}\to \RR$ such that
\ba
	\mathrm{MMD}^2[\mathscr{F},p,q] & = 
	\EE_{x,x'\sim p}[k(x,x')] 
	- 2 \EE_{x\sim p, y\sim q} [k(x,y)] \notag
	\\
	& \quad + \EE_{y,y' \sim q} [k(y,y')]\,.
	\label{eq:2121}
\ea
MMD has been recently used in deep reinforcement learning to model the distribution of future returns \cite{Tang2021,Zhang2021x}.

A closely related quantity that measures the statistical distance between two distributions in Euclidean space is the so-called \emph{energy distance} \cite{Szekely2003,Szekely2004x,Szekely2013,Szekely2017}, defined as
\ba
	\mathcal{D}_E(p,q) & := 2 \EE_{\x\sim p, \y\sim q} \| \x-\y \| 
	- \EE_{\x,\x'\sim p} \|\x-\x' \| 
	\notag
	\\
	& \quad ~ - \EE_{\y,\y'\sim q} \| \y - \y' \|
	\label{eq:gfdfg769}
\ea
where $\|\cdot \|$ stands for the Euclidean $L_2$ norm. 
Equation~\eqref{eq:gfdfg769} bears close similarity to \eqref{eq:2121}. In fact, under suitable conditions, they are proven to be equivalent \cite{Sejdinovic2013,Shen2018}. 

A useful finite-sample version of \eqref{eq:gfdfg769} is given by
\ba
	\mathcal{D}_E(p,q) & = \frac{2}{mn}\sum_{i=1}^m
	\sum_{j=1}^n \|\x_i - \y_j \| - \frac{1}{m^2}\sum_{i=1}^m
	\sum_{j=1}^m \|\x_i - \x_j \| \notag
	\\
	& \quad 
	- \frac{1}{n^2} \sum_{i=1}^n \sum_{j=1}^n \| \y_i - \y_j \|\,.
	\label{eq:f3229}
\ea

\subsection{How to Measure the Reliability of a Probabilistic Prediction?}

Uncertainty of a prediction can be expressed in miscellaneous ways. Confidence intervals are a popular and simple example, whereas the shape of the probability density can also be specified. How to assess the reliability of such forecasts? If we had the knowledge of the data-generating process and knew the exact form of the true conditional probability $p(\y|\x)$, it would be straightforward to assess the accuracy of a probabilistic forecast $p(\wh{\y}|\x)$ by using distance metrics such as MMD and the energy distance; however, this is not possible in general, as we only have access to a single realization $\{(\x_i, \y_i)\}_{i=1}^N$ of the data-generating process. This issue was investigated in detail by \cite{Gneiting2007}, who introduced the important concept of \emph{proper scoring rules}. When the response variable $\y$ is a scalar, a widely used \emph{strictly proper} scoring rule for probabilistic forecast is the continuous ranked probability score (CRPS) defined as 
\ba
	\mathrm{CRPS}(F, y) & :=\int_{-\infty}^{\infty}\rmd \wh{y} \; 
	\kkakko{F(\wh{y}) - \1\{\wh{y}\geq y\}}^2 \,,
\ea
where $F$ is the cumulative distribution function of a prediction, and $\1\{\diamondsuit\}:=1$ if $\diamondsuit$ is true and $=0$ otherwise. When the prediction is deterministic (i.e., a point forecast), CRPS coincides with the mean absolute error (MAE), so CRPS can be considered as a probabilistic generalization of MAE.  Interestingly, CRPS may be cast into the form \cite{Gneiting2007}
\ba
	\mathrm{CRPS}(F, y) & = \EE_{F}|\wh{y}-y| 
	- \frac{1}{2}\EE_{F}|\wh{y}-\wh{y}'| \,,
	\label{eq:fgd8}
\ea
where $\wh{y}$ and $\wh{y}'$ are independent copies of a random variable with the cumulative distribution function $F$ and $\EE_{F}$ is the expectation value with regard to $F$. Note that \eqref{eq:fgd8} agrees with $1/2$ of the energy distance \eqref{eq:gfdfg769} with a single sample of $y$. It is important that \eqref{eq:fgd8} is \emph{strictly proper}, namely, its expectation value w.r.t.~the distribution of $y$, i.e.~$\EE_y[\mathrm{CRPS}(F, y)]$, is minimized if and only if $F$ coincides with the true cumulative distribution of $y$. 

The expression \eqref{eq:fgd8} readily lends itself to a multi-dimensional generalization
\ba
	\mathrm{ES}(P, \y) = \frac{1}{2}\EE_{\wh{y},\wh{y}'\sim P} \|\wh{\y} - \wh{\y}'\|^\alpha - \EE_{\wh{y}\sim P} \|\wh{\y}-\y\|^\alpha \,,
	\label{eq:dfsgfd4}
\ea
which is called the \emph{energy score} \cite{Gneiting2008,Pinson2012} and has been heavily used in meteorology. (Note that \emph{lower} CRPS is better and \emph{higher} energy score is better.) The energy score is strictly proper for $0<\alpha<2$  \cite{Szekely2003}.

\section{\uppercase{Generative ensemble prediction based on energy scores}\label{sc:89dsuf}}

In this section, we give an overview of DISCO Nets \cite{disco2016}, which make a density forecast based on sample generation. 

The first step in this method is to enlarge the feature space, from the original one $\mathscr{X}$ to $\mathscr{X}\times\Xb$, where $\Xb$ is an arbitrary base space. The dimension of $\Xb$ must be greater than or equal to the dimension of the target variable $\y$. A convenient choice for $\Xb$ would be $[0,1]^d$. The next step is to select a base distribution $\Pb$ over $\Xb$, which may be simply a uniform distribution. The training of NN proceeds along a standard stochastic gradient descent style. We take a minibatch of samples from the training dataset and ``augment'' every sample with a random vector sampled from $\Pb$. That is, each input vector $\x_i$ is first duplicated $\Nb$ times, and then each copy is paired with an independently sampled random vector from $\Xb$. As a result, the minibatch size increases from $b_{\rm batch}$ to $b_{\rm batch} \times \Nb$. The resulting ``elongated'' input vectors are fed into the NN and the outputs $\big\{\wh{\y}_i^{(n)}\big\}_{n=1}^{\Nb}$ are obtained. Finally, the loss function $\L$ is computed by using the true regression target $\y_i$ and the model predictions $\big\{\wh{\y}_i^{(n)}\big\}_{n=1}^{\Nb}$ according to the formula 
\ba
	\L \mkakko{\y, \big\{\wh{\y}^{(n)}\big\}_{n=1}^{\Nb}} & := \frac{1}{\Nb}
	\sum_{i=1}^{\Nb}\| \y-\wh{\y}^{(i)} \| 
	\notag \\
	& \quad \; 
	- \frac{1}{2\Nb^2}\sum_{i=1}^{\Nb}\sum_{j\ne i}
	\|\wh{\y}^{(i)} - \wh{\y}^{(j)}\| \,.
	\label{eq:o93cd}
\ea
This is a negated finite-sample energy score \eqref{eq:dfsgfd4} with $\alpha=1$. The first term of \eqref{eq:o93cd} encourages all samples $\wh{\y}^{(i)}$ to come closer to $\y$, whereas the second term of \eqref{eq:o93cd} induces a repulsive force between samples so that the samples' cloud swells. This loss function is summed across all data in the minibatch and the NN parameters are updated in the direction of the gradient descent of the loss. Since the energy score is a \emph{strictly proper} scoring rule, its expectation value is maximized if and only if the evaluated density forecast agrees with the true density of $\y$. Hence it can be expected that the above training will let the NN predictions converge to the true data-generating distribution. For a formal justification on this point, see Appendix~\ref{ap:9hufg}. To the best of our knowledge, the elementary argument presented in Appendix~\ref{ap:9hufg} is new. 

In the test phase, for a given input vector $\x$, we can generate as many ensemble predictions as we like from the NN by first duplicating $\x$, then augmenting each copy with an independent random vector from $\Pb$, and finally feeding them into the NN. Statistical quantities such as the mean, standard deviation and quantile points can be readily computed from the resulting ensemble of predictions. 

DISCO Net (and {\ALG}) differs from recent approaches based on NF \cite{Sick2020,Charpentier2020,Dumas2021,Sendera2021,Jamgochian2022,Rittler2022,Marz2022,Arpogaus2022,Cramer2022} in a number of ways. First, NF estimates the joint density $p(\x,\y)$ to derive the conditional density $p(\y|\x)$ while DISCO Net directly models the latter with no recourse to the former. Second, NF maximizes the log-likelihood to optimize parameters while DISCO Net does not use the log-likelihood at all during training. Third, and related to the second point, NF needs a differentiable and \emph{invertible} mapping, whereas DISCO Net is free from such restrictions, which leads to more flexible modeling and considerable simplification of implementation. 

When compared with Monte Carlo dropout \cite{Gal2016}, which is one of the most popular UQ methods, the main difference is that the NN in DISCO Net/{\ALG} is deterministic and no probabilistic sampling of network structures is performed. 

It is worthwhile to note that the size of the prediction ensemble in DISCO Net/{\ALG} is arbitrary, and can be increased at no extra cost in the test phase, in contrast to the existing methods \cite{Tang2021,Zhang2021x} that have a fixed number of network outputs to model the ensemble. 

Finally, we stress that DISCO Net/{\ALG} is nonparametric and, in principle, can model an arbitrary distribution in arbitrary dimensions, whereas networks that learn quantile points of a distribution using a quantile loss \cite{QRDQN2018,IQN2018,Yang2019,Singh2022} generally fail to model a distribution in more than one dimension.

\section{\uppercase{Experiments}\label{sc:hgf98dfbv}}
In this section we introduce improvements to DISCO Nets and conduct numerical experiments on synthetic and real tabular datasets to assess the reliability of our ensemble forecasts.

\subsection{Evaluation Metrics\label{sc:emet}}
When the true underlying data-generating distribution is known, we can use the following metrics to directly measure the discrepancy between the predicted density and the true density.
\begin{itemize}[leftmargin=10pt]
	\item The Jensen--Shannon distance (JSD) \cite{Endres2003}:~
	This quantity is the square root of the Jensen--Shannon divergence (JSdiv) defined as
	\ba
		\hspace{-2mm}
		\mathrm{JSD}(P\|Q)^2 & = 
		\mathrm{JSdiv}(P\|Q) 
		\\
		&:= \frac{1}{2}D_{\rm KL}(P\|M) + 
		\frac{1}{2}D_{\rm KL}(Q\|M)\,, \hspace{-2mm}
	\ea
	where $D_{\rm KL}(P\|Q)$ is the Kullback–Leibler divergence \cite{KL1951} between the distributions $P$ and $Q$, and $M:=(P+Q)/2$. When the logarithm with base 2 is used for computation, $0\leq \mathrm{JSD}(P\|Q)\leq 1$ holds.
	\item The Hellinger distance \cite{EMS2001}:~
	This quantity is given in terms of probability densities as
	\ba
		\hspace{-2mm}
		\mathrm{HD}(P,Q) & := \sqrt{\frac{1}{2}\int \rmd \x 
		\mkakko{\sqrt{p(\x)} - \sqrt{q(\x)}}^2}\,.
	\ea
	It satisfies the bound $0 \leq \mathrm{HD}(P,Q) \leq 1$.
	\item The first Wasserstein distance:~In terms of the cumulative distribution functions $F_p(x)$ and $F_q(x)$ of $P$ and $Q$, we have, for one-dimensional distributions,
	\ba
		\mathrm{WD}(P,Q)=\int_{-\infty}^{\infty}\rmd x\; |F_p(x) - F_q(x)|\;.
	\ea
	\item The square root of the energy distance \eqref{eq:gfdfg769}:~Similarly to the Wasserstein distance, in terms of one-dimensional cumulative distribution functions we have \cite{Szekely2003}\footnote{The square root is taken here to match the definition of the energy distance function in Scipy: \url{https://docs.scipy.org/doc/scipy/reference/generated/scipy.stats.energy_distance.html}.}
	\ba
		\mathrm{ED}(P,Q) &:= \sqrt{\mathcal{D}_E(P,Q)} 
		\\
		& = \sqrt{2\int_{-\infty}^{\infty}\rmd x\; \mkakko{F_p(x) - F_q(x)}^2}\;.
	\ea
\end{itemize}

\subsection{Neural Network Architecture\label{sc:u90gbf}}

\begin{figure}
	\centering
	\includegraphics[height=5cm]{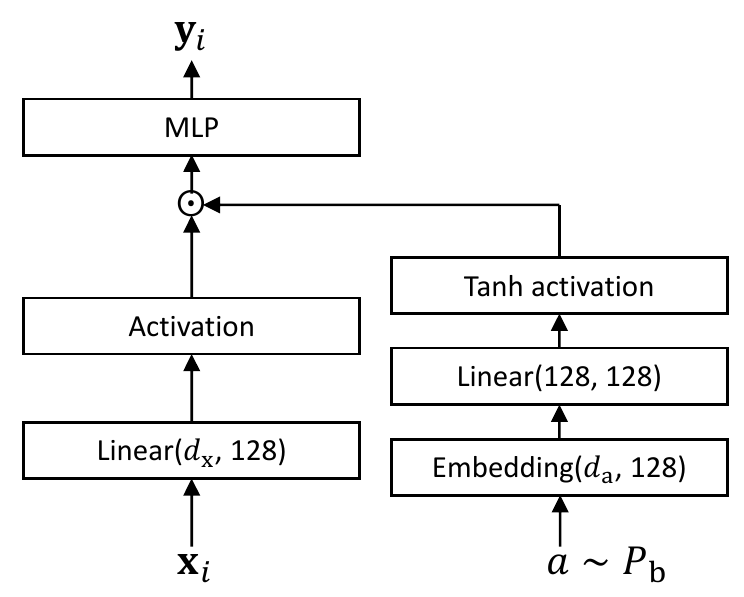}
	\caption{\label{fg:43rds}Architecture of a NN used to make distributional predictions by {\ALG}. Here $d_{\x}$ is the feature dimension of $\x$ and $d_{\rm a}$ is the dimension of $a$.}
\end{figure}

When the response variable $\y$ to be predicted is a scalar, we adopt a NN architecture depicted in figure~\ref{fg:43rds} with $d_{\rm a}=1$. 
Inspired by \cite{IQN2018} we convert a scalar $a$ sampled from a uniform distribution over $[0,1]$ to a 128-dimensional vector $\mkakko{1, \cos(\pi a), \cos(2 \pi a), \cdots, \cos(127 \pi a)}$, which is later merged into the layer of $\x$ via a dot product. For the MLP part of the network, we used 2 hidden layers of width 128, each followed by an activation layer. The network parameters are optimized with Adam. 

When $\y$ is a two-dimensional vector, we draw a random vector $\a=(a_1,a_2)$ from the uniform distribution over $[0,1]^2$ and expand it to a 128-dimensional vector $\mkakko{1, \cos(\pi a_1), \cos(2 \pi a_1), \cdots, \cos(63 \pi a_1)}\oplus \mkakko{1, \cos(\pi a_2), \cos(2 \pi a_2), \cdots, \cos(63 \pi a_2)}$.

The activation function for the embedding layer is changed from ReLU in \cite{IQN2018} to Tanh. This choice was partly motivated by \cite[Appendix B.3]{Ma2020}, which recommended using Sigmoid instead of ReLU. On the other hand, activation functions for the MLP part are not fixed and can be changed flexibly. 

As we will see later, this surprisingly simply trick of inserting an embedding layer of an external noise works perfectly well for stable and efficient NN training in {\ALG}. 

\subsection{Test on a Unimodal Synthetic Dataset}

We tested the proposed method on a simple one-dimensional dataset (figure~\ref{fg:24fds}, left). The data comprising 200 points were generated from the distribution
\begin{gather}
	y = \exp(\sin(\pi x)) + \epsilon, \label{eq:rt34b52}
	\\
	\epsilon \sim \mathcal{N}\mkakko{0, 0.4^2}, \quad -1\leq x \leq 1\,.
\end{gather}
We applied {\ALG} to this data with the hyperparameters $n_{\rm epoch}=300, \Nb=50$ and $b_{\rm batch}=20$, using Tanh layer for the activation function. All parameters of the linear layers were initialized with random variables drawn from $\mathcal{N}\mkakko{0,0.1^2}$. The obtained distributional forecast (figure~\ref{fg:24fds}, right) captures aleatoric uncertainty of the data reasonably well. The color in the figure represents $0.5 - |Q - 0.5|$, where $Q\in [0,1]$ denotes the quantile of the predictive density.

To quantitatively assess the result, we took 200 equidistant points over the interval $[-1,1]$ of the $x$-axis and, for each $x$ in this set, sampled 3000 points from the ensemble prediction of {\ALG}. Then we computed the discrepancy between the obtained distribution and the ground-truth distribution \eqref{eq:rt34b52} and took the average of the distance metric over all 200 points. 

For comparison, we performed similar analyses with three popular UQ methods: LightGBM with a quantile loss \cite{lightgbm}, quantile regression forests (QRF) \cite{Meinshausen2006}, and Gaussian process regression (GP) \cite{RWbook}. For QRF and GP we have used the implementation of \cite{Scikit-Garden} and \cite{scikit-learn}, respectively. Hyperparameters of LightGBM and QRF were tuned with Optuna \cite{Akiba2019}. We computed 51 quantiles with LightGBM and QRF, and therefrom constructed the conditional probability density $p(y|x)$ approximately. The result was so spiky that we had to smooth it with a Gaussian filter. 

\begin{figure}
	\centering
	\includegraphics[width=5cm]{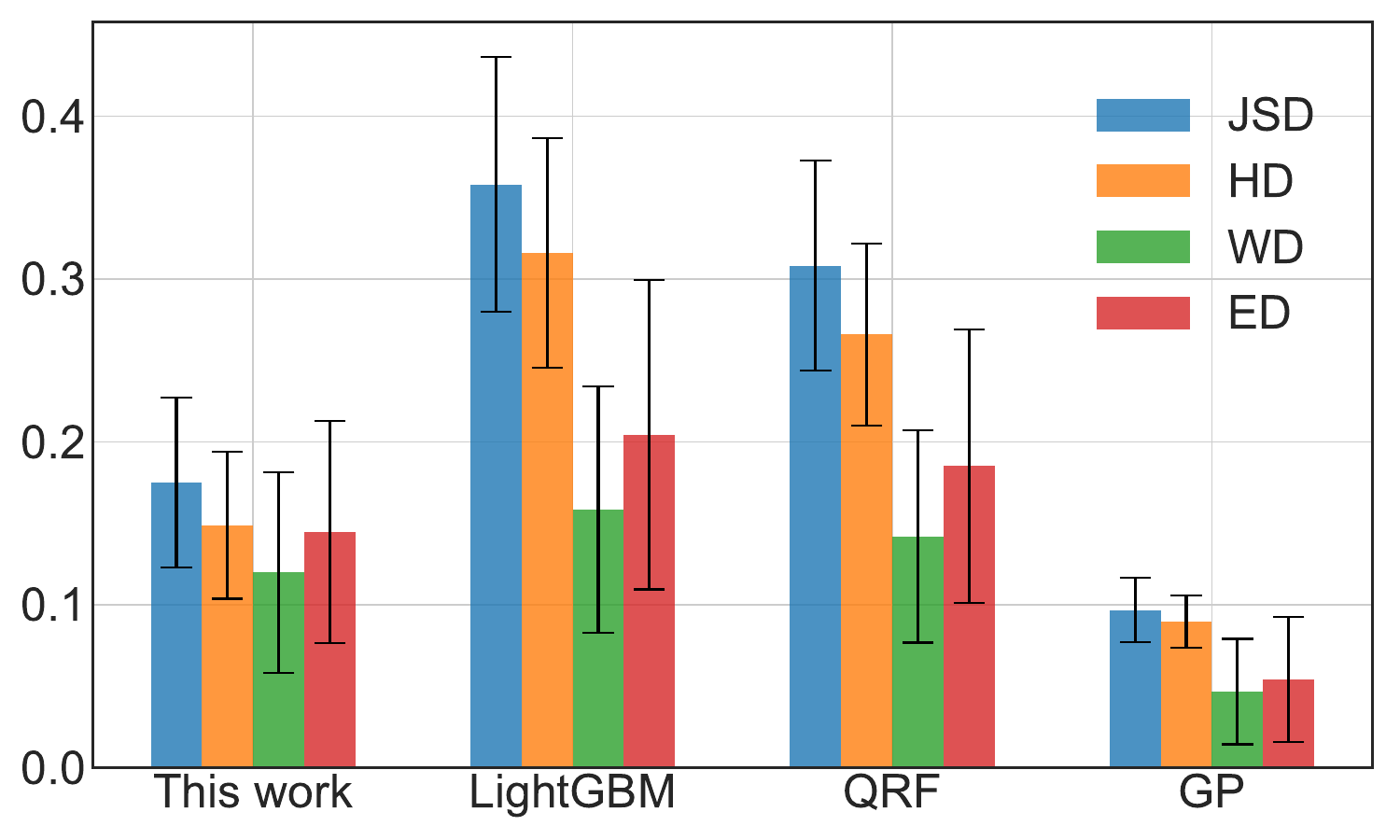}
	\caption{\label{fg:8h9gf324}Average distance between the true distribution $p(y|x)$ and the output of each distributional predictors. The error bars represent the mean $\pm$ one standard deviation.}
\end{figure}

The benchmark result is summarized in figure~\ref{fg:8h9gf324}. (For the definition of four distance metrics, we refer to section~\ref{sc:emet}.) The result clearly shows that {\ALG} yields better distributional forecasts than LightGBM and QRF. However, {\ALG} is outperformed by GP, which does not come as a surprise because the ground-truth distribution \eqref{eq:rt34b52} is Gaussian. 
We once again emphasize that the result for {\ALG} has been obtained in a fully nonparametric manner without assuming any parametric distribution such as Gaussian.

\subsection{Test on a Multimodal Synthetic Dataset}

\begin{figure}
	\centering
	\includegraphics[width=5cm]{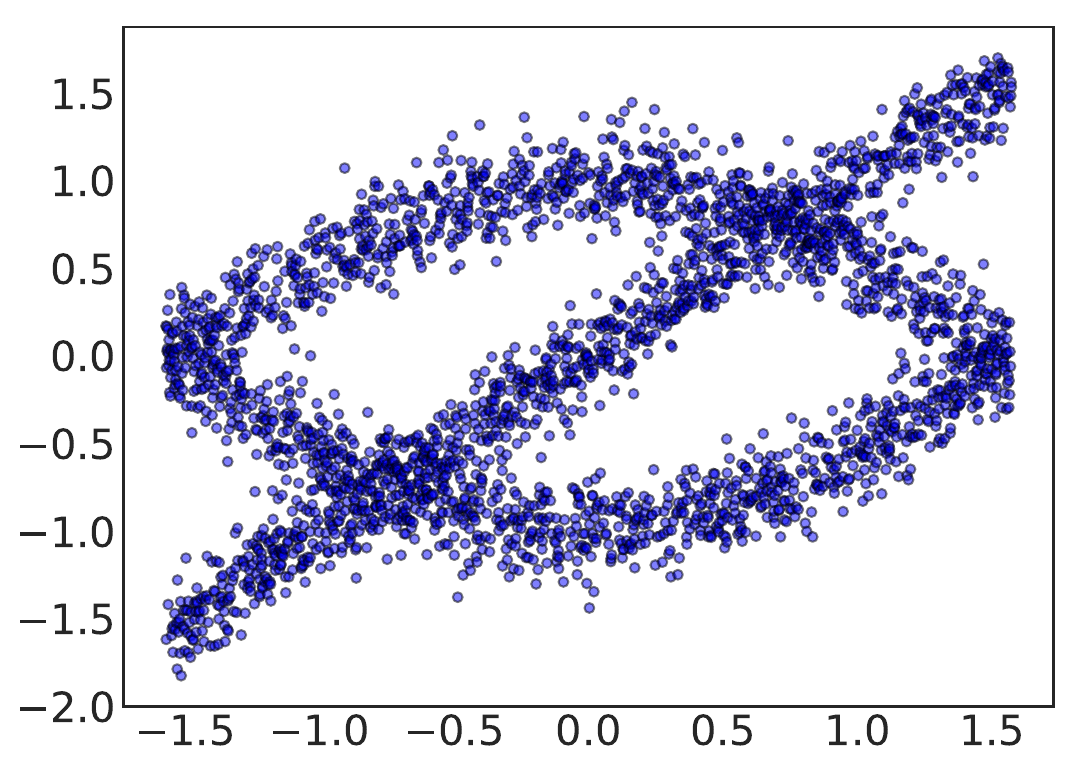}
	\caption{\label{fg:645kojk}Data comprising 3000 points generated by adding a Gaussian noise to three crossing curves.}
	\vspace{2mm}
	\centering
	\includegraphics[width=5cm]{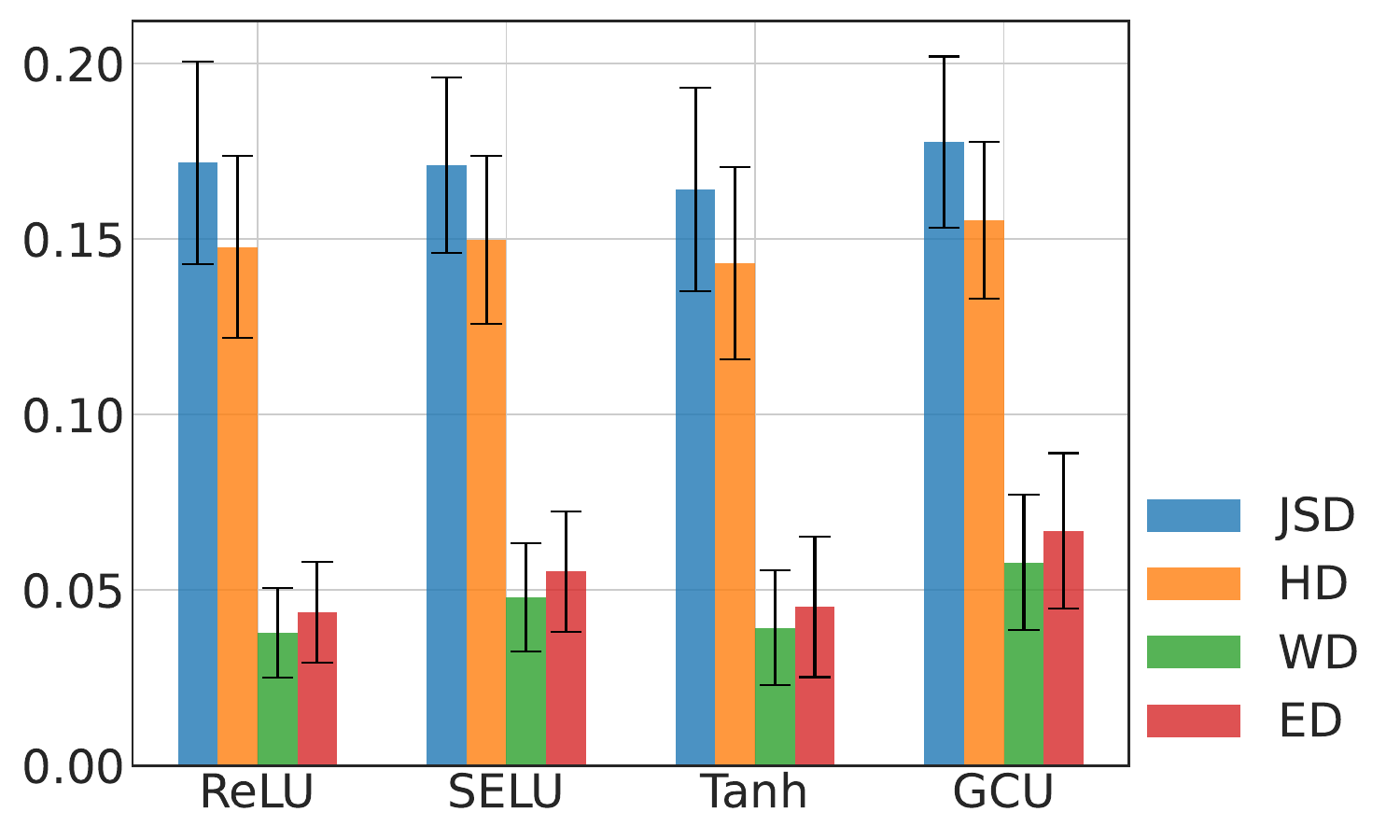}
	\caption{\label{fg:5jkh}Average distance between the predictive distribution and the true distribution after training with four kinds of activation functions. The error bars are one standard deviation above and below the mean.}
\end{figure}

Next, we test the proposed method on a multimodal synthetic dataset. 
The data were generated from the distributions over $-\pi/2\leq x \leq \pi/2$,
\ba
	\begin{cases}
		y=x+\varepsilon, \\
		y=\cos x + \varepsilon',\\ 
		y=-\cos x + \varepsilon'',
	\end{cases} 
	\varepsilon,\varepsilon',\varepsilon''\sim \mathcal{N}(0,0.15^2).
	\label{eq:46541236}
\ea
The scatter plot of the data is shown in figure~\ref{fg:645kojk}. The dataset includes 3000 points in total, 1000 for each distribution. 
As a function of $x$, $y$ is multimodal and in general there are three peaks in the conditional probability density $p(y|x)$.

To test the effectiveness of {\ALG}, we trained a NN with the hyperparameters $n_{\rm epoch}=300$, $\Nb=50$ and $b_{\rm batch}=150$. We conducted training with four distinct activation functions: Rectified Linear Unit (ReLU), Scaled Exponential Linear Unit (SELU) \cite{Klambauer2017}, Tanh activation, and Growing Cosine Unit (GCU) \cite{Noel2021}. To measure the performance, we took 100 equidistant points $x\in[-\pi/2, \pi/2]$ and, for each of them, computed the estimate $p(\wh{y}|x)$ via $M=10^4$ ensemble prediction. The distance between $p(\wh{y}|x)$ and the true conditional $p(y|x)$ was measured with the metrics in section~\ref{sc:emet} and was then averaged over all 100 points. The result is shown in figure~\ref{fg:5jkh}. It is observed that the scores of all four activation functions are similar, although Tanh activation seems to perform slightly better than the others. 

\begin{figure}
	\centering
	\includegraphics[width=.47\textwidth]{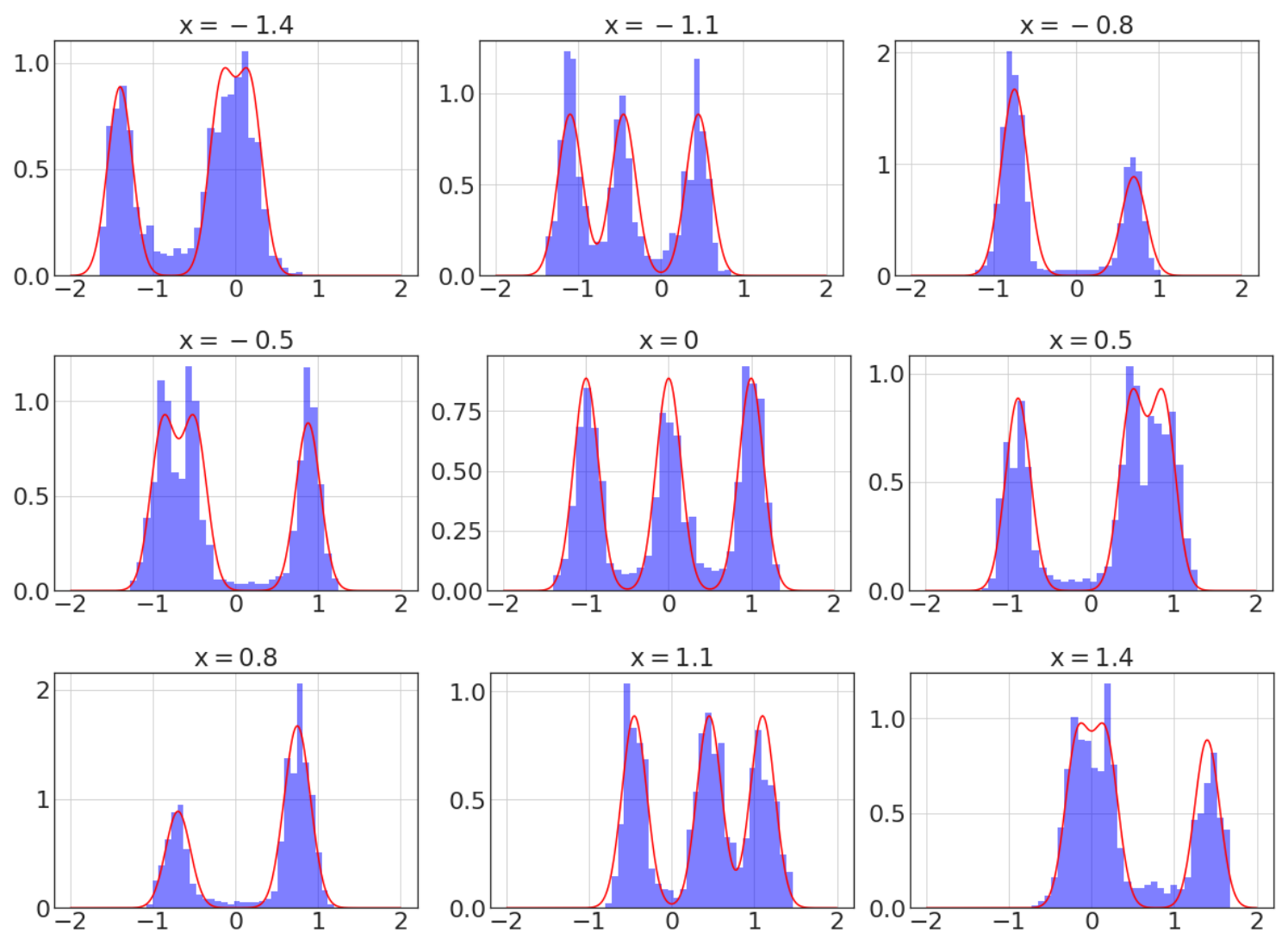}
	\caption{\label{fg:moiijjh}The histogram of the predicted conditional distribution $p(\wh{y}|x)$ (blue) overlayed with the true distribution $p(y|x)$ (red) for the dataset in figure~\ref{fg:645kojk}. Tanh activation was used.}
\end{figure}

To gain more insights into the approximation quality of our method, we show the predicted distribution versus the ground truth for various values of $x$ in figure~\ref{fg:moiijjh}. Clearly the predicted distribution reproduces the complex multimodal shape of the true distribution in a satisfactory manner. This is not possible with conventional methods such as the Gaussian processes. 

We have also applied plain vanilla DISCO Nets (with no embedding layer of a random noise) to this dataset, but it only yielded seriously corrupted estimates of the predictive distribution. Actually it was this failure that forced us to seriously consider modifications to DISCO Nets. 

\begin{figure}
	\centering
	\includegraphics[width=.234\textwidth]{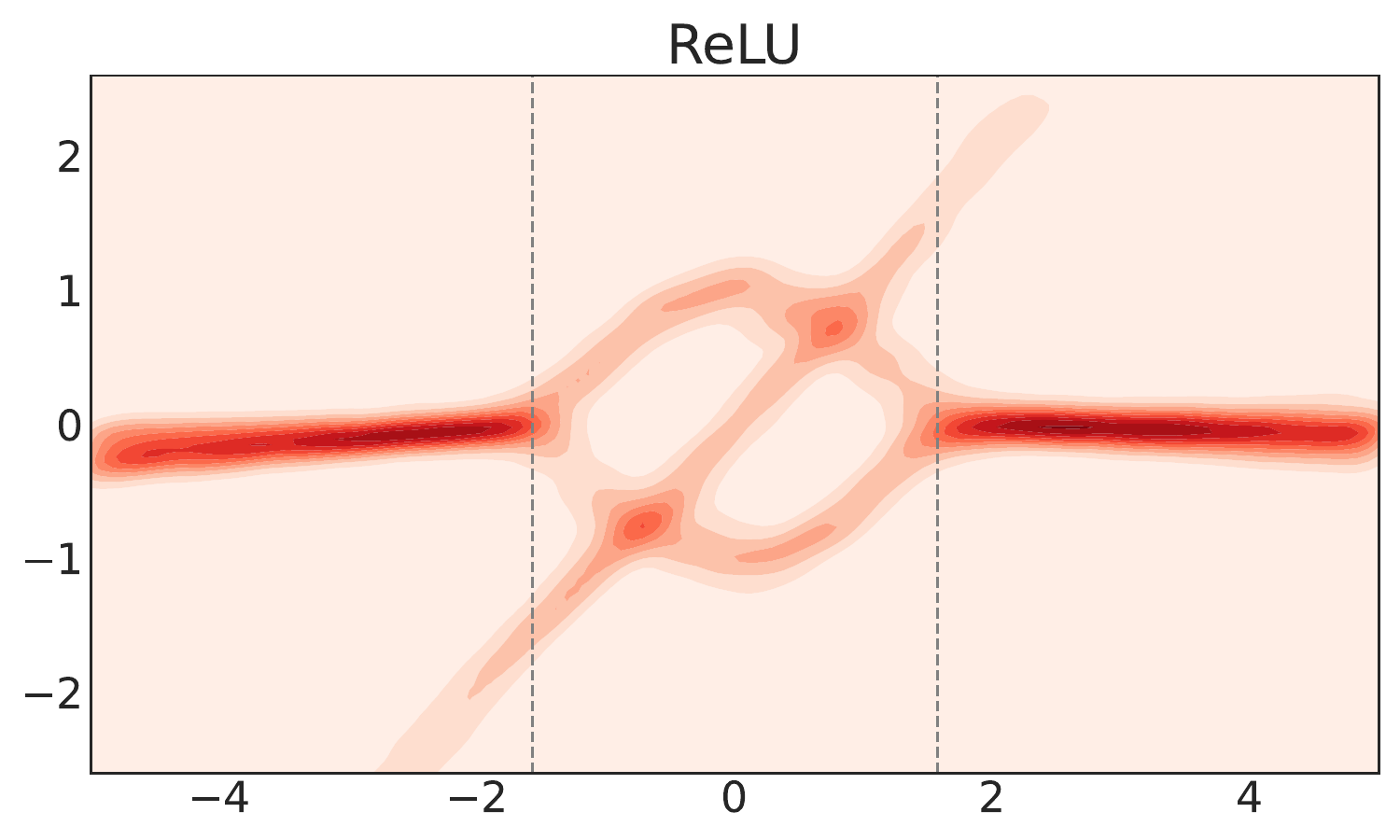}
	\includegraphics[width=.234\textwidth]{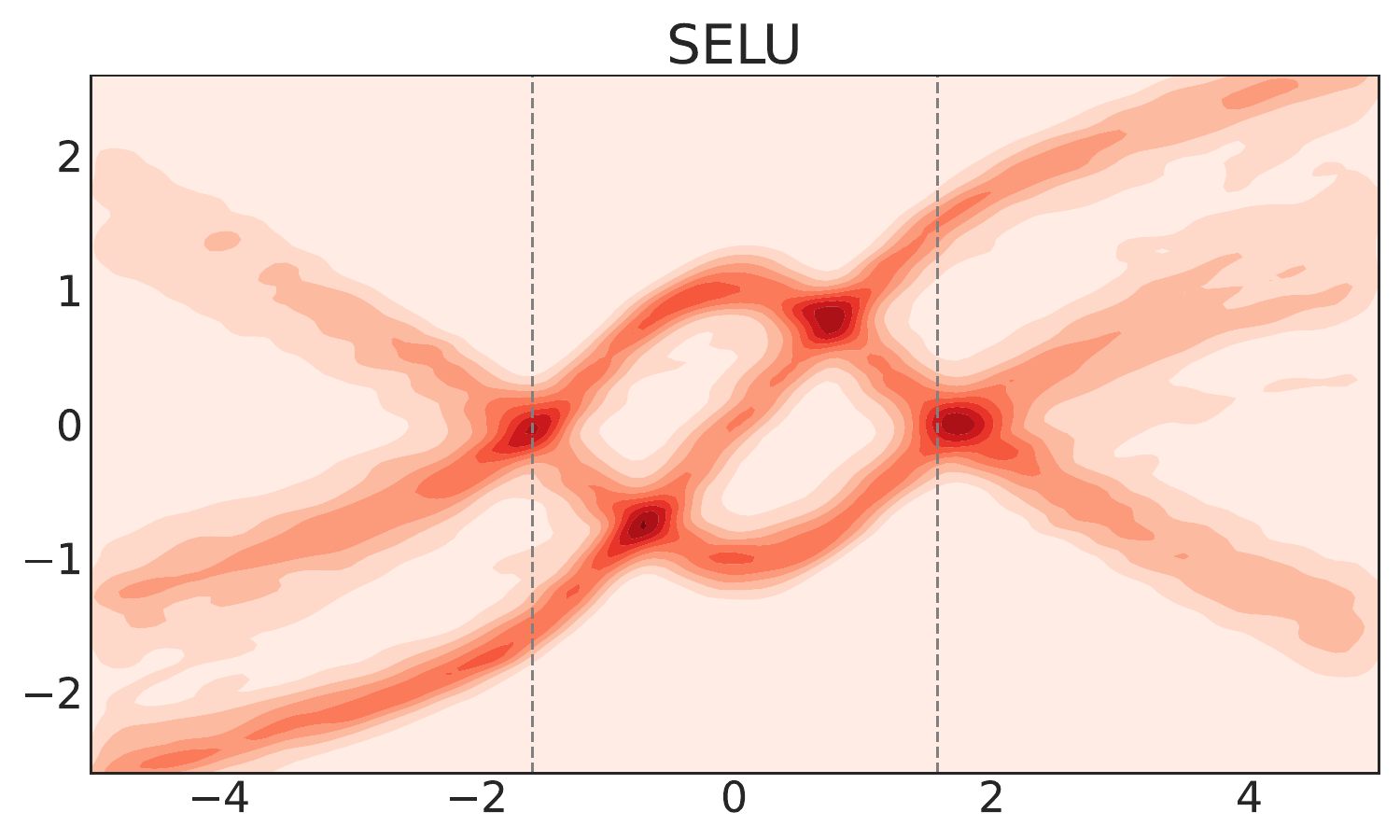}
	\\
	\includegraphics[width=.234\textwidth]{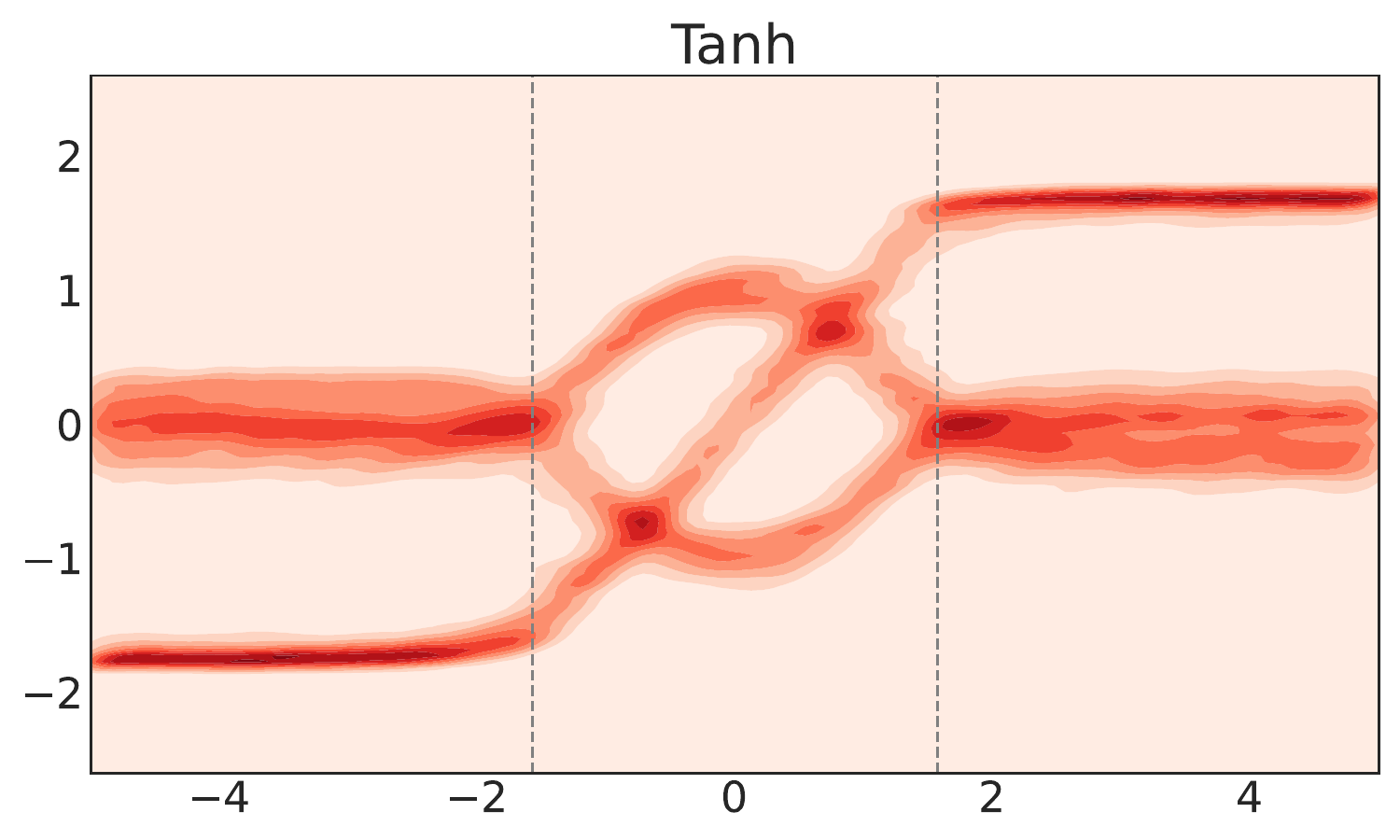}
	\includegraphics[width=.234\textwidth]{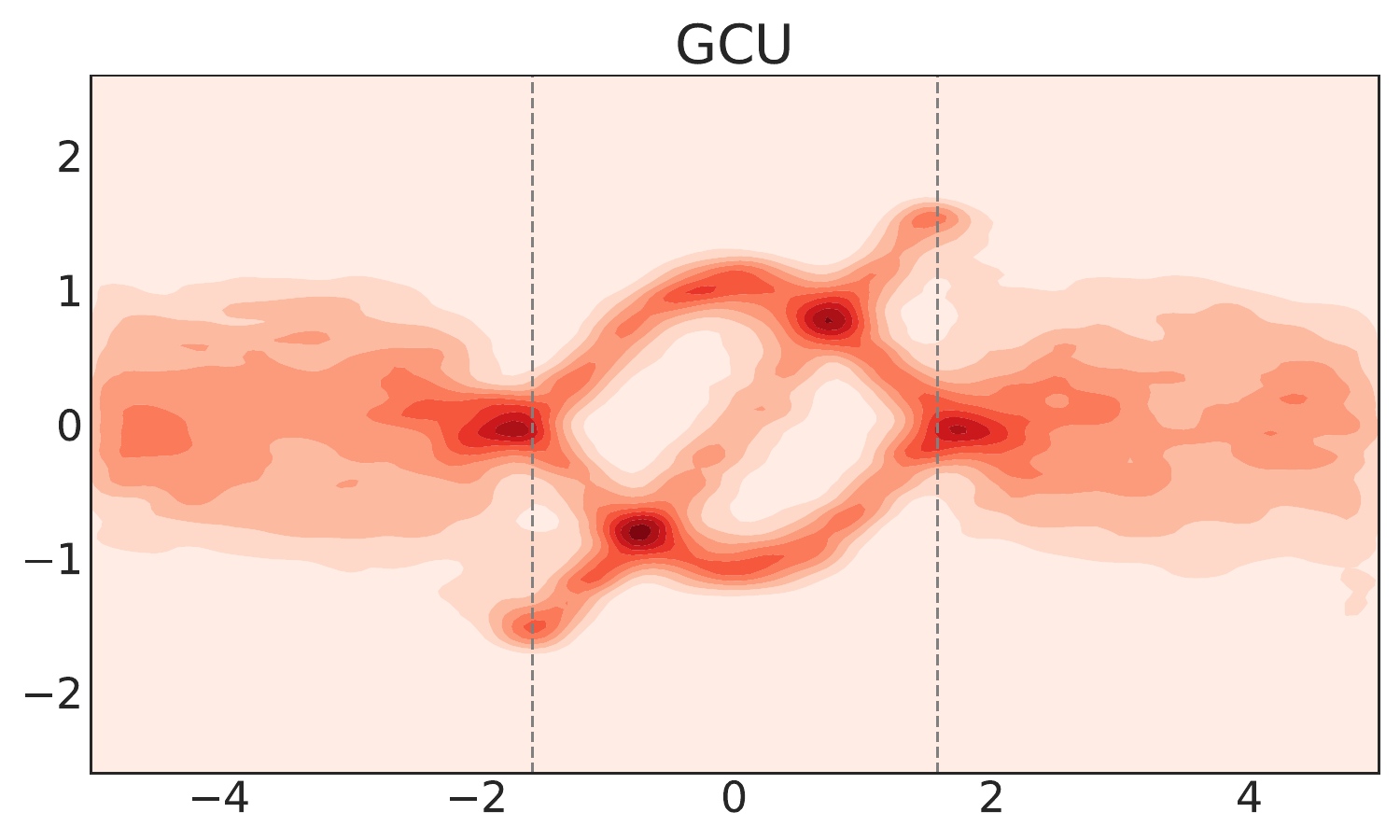}
	\caption{\label{fg:798gho}Density plots of the predictive distributions for $-5\leq x \leq 5$ with four activation functions. Two vertical dashed lines at $x=\pm \pi / 2$ indicate the boundary of the training dataset.}
\end{figure}

While all activation functions give comparable accuracy of approximation for the range of data $-\pi/2\leq x \leq \pi /2$, what happens outside this range? Do they still maintain similar functional forms? We have numerically studied this and found, as shown in figure \ref{fg:798gho}, that the four neural networks give drastically different extrapolations. Since no data is available for $|x|>\pi/2$, the epistemic uncertainty must be high and a predictive distribution of $y$ should have a broad support. This seems to hold true only for the network with GCU activation. We therefore conclude that, while the choice of an activation function hardly affects the network's capability to model aleatoric uncertainty for in-distribution data, it \emph{does} affect the ability to model epistemic uncertainty for out-of-distribution data and hence great care must be taken.

\begin{figure}
	\centering
	\includegraphics[width=.485\textwidth]{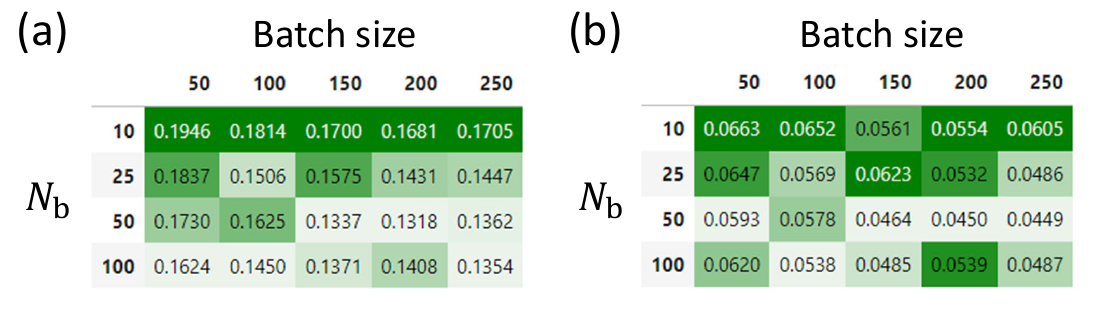}
	\caption{\label{fg:hodfkho}(a) Hellinger distance and (b) Wasserstein distance between the true distribution $p(y|x)$ and the predictive distribution $\wh{p}(y|x)$ computed using ReLU activation for varying $b_{\rm batch}$ and $\Nb$.}
\end{figure}

To check the hyperparameter dependence of results, we have repeated numerical experiments with the same dataset for various sets of $b_{\rm batch}$ and $\Nb$, using ReLU activation. The number of epochs $n_{\rm epoch}$ was fixed to $2 \times b_{\rm batch}$ to keep the number of gradient updates the same across experiments, ensuring a fair comparison. The result is shown in figure~\ref{fg:hodfkho}. It is observed that larger $\Nb$ and/or larger $b_{\rm batch}$ generally leads to better (lower) scores. Note, however, that larger $\Nb$ entails higher numerical cost and longer computational time. It is therefore advisable to make these parameters large enough within the limit of available computational resources, although it is likely that the best hyperparameter values will in general depend on the characteristics of individual datasets under consideration. 

In order to benchmark {\ALG}, we again compared it with LightGBM, QRF and GP.  
\begin{figure}
	\centering
	\includegraphics[width=.48\textwidth]{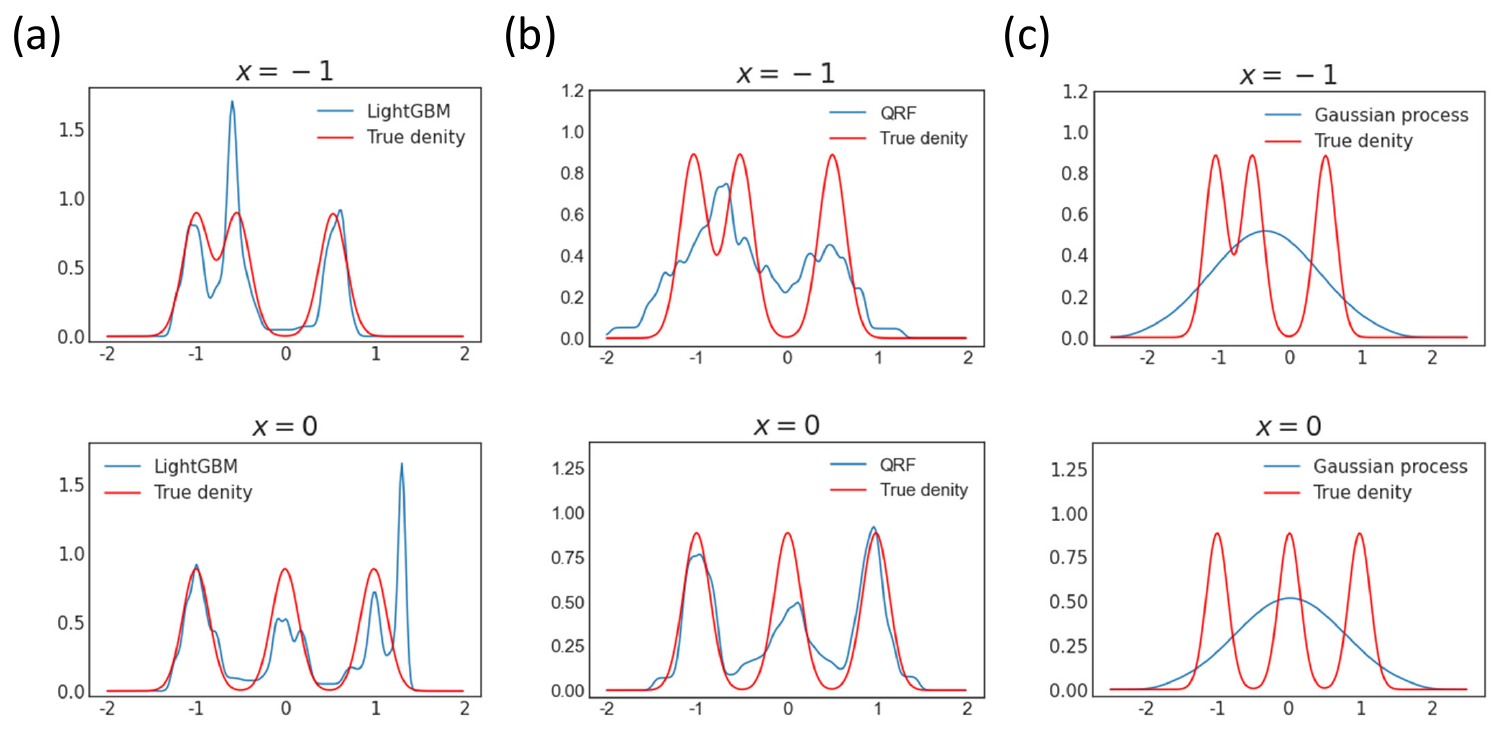}
	\caption{\label{fg:54po54}The true density $p(y|x)$ (red curve) and the estimated density $\wh{p}(y|x)$ (blue curve) obtained at $x=-1$ and $0$ for the dataset in figure~\ref{fg:645kojk} using the three methods: (a) LightGBM with quantile losses, (b) QRF, and (c) GP.}
\end{figure}
We have used ReLU activation for {\ALG}. Some examples of the posterior distributions obtained with each method are displayed in figure~\ref{fg:54po54}. While LightGBM and QRF seem to capture qualitative features of the posterior well, GP entirely misses local structures due to its inherent Gaussian nature of the approximation. 
\begin{figure}
	\centering
	\includegraphics[width=.42\textwidth]{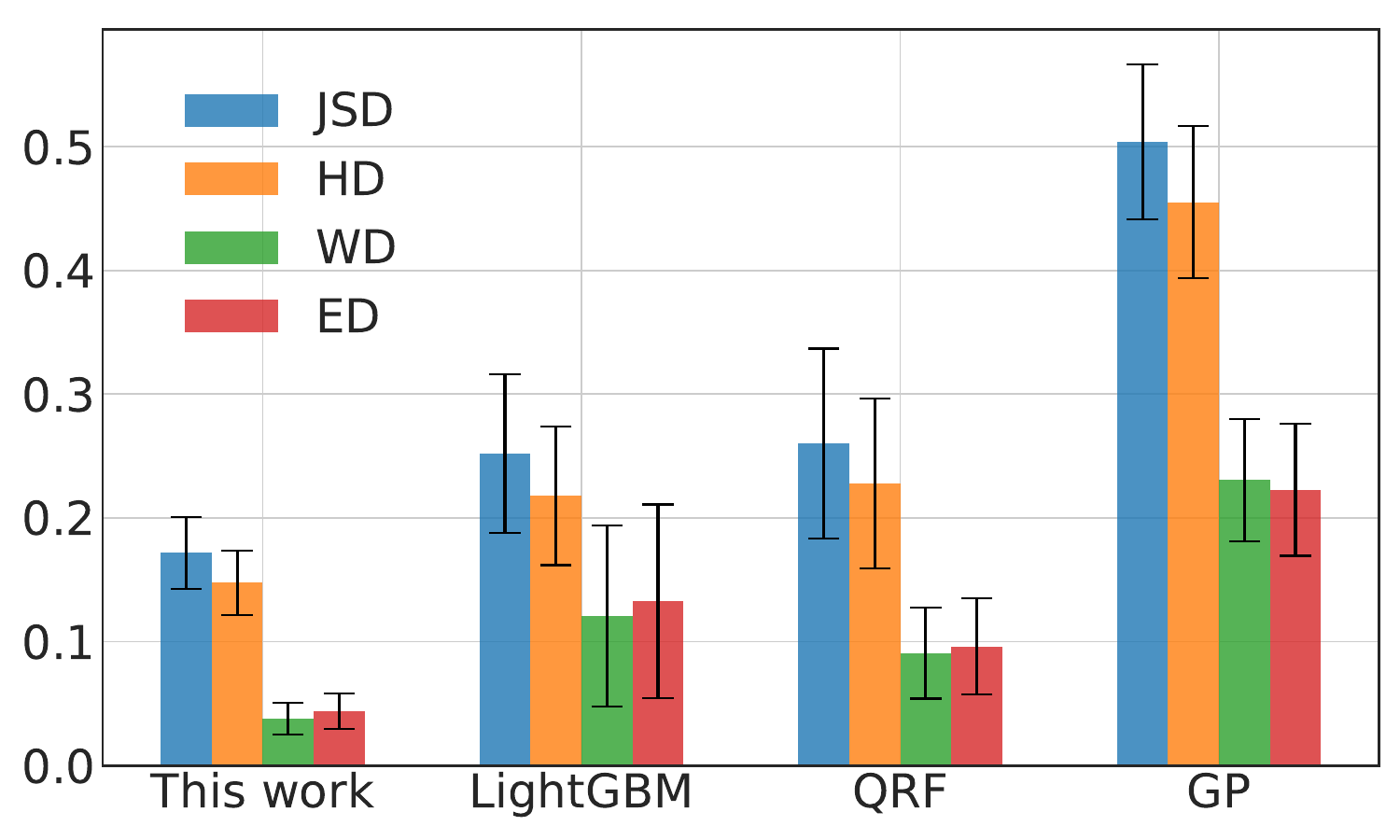}
	\caption{\label{fg:7ghfhjh}Comparison of {\ALG} against three popular UQ methods (lower scores are better). Error bars represent the mean $\pm$ one standard deviation.}
\end{figure}

For a quantitative comparison of the methods, we computed the discrepancy between the estimated distribution and the true distribution and took the average over $-\pi/2 \leq x \leq \pi/2$. The result is displayed in figure~\ref{fg:7ghfhjh}. Not surprisingly, GP marked the worst score. {\ALG} attained the lowest score in all the four metrics, thus underlining the effectiveness of this approach for modeling complex multimodal probability distributions. 

\begin{figure}
	\centering
	\includegraphics[width=.48\textwidth]{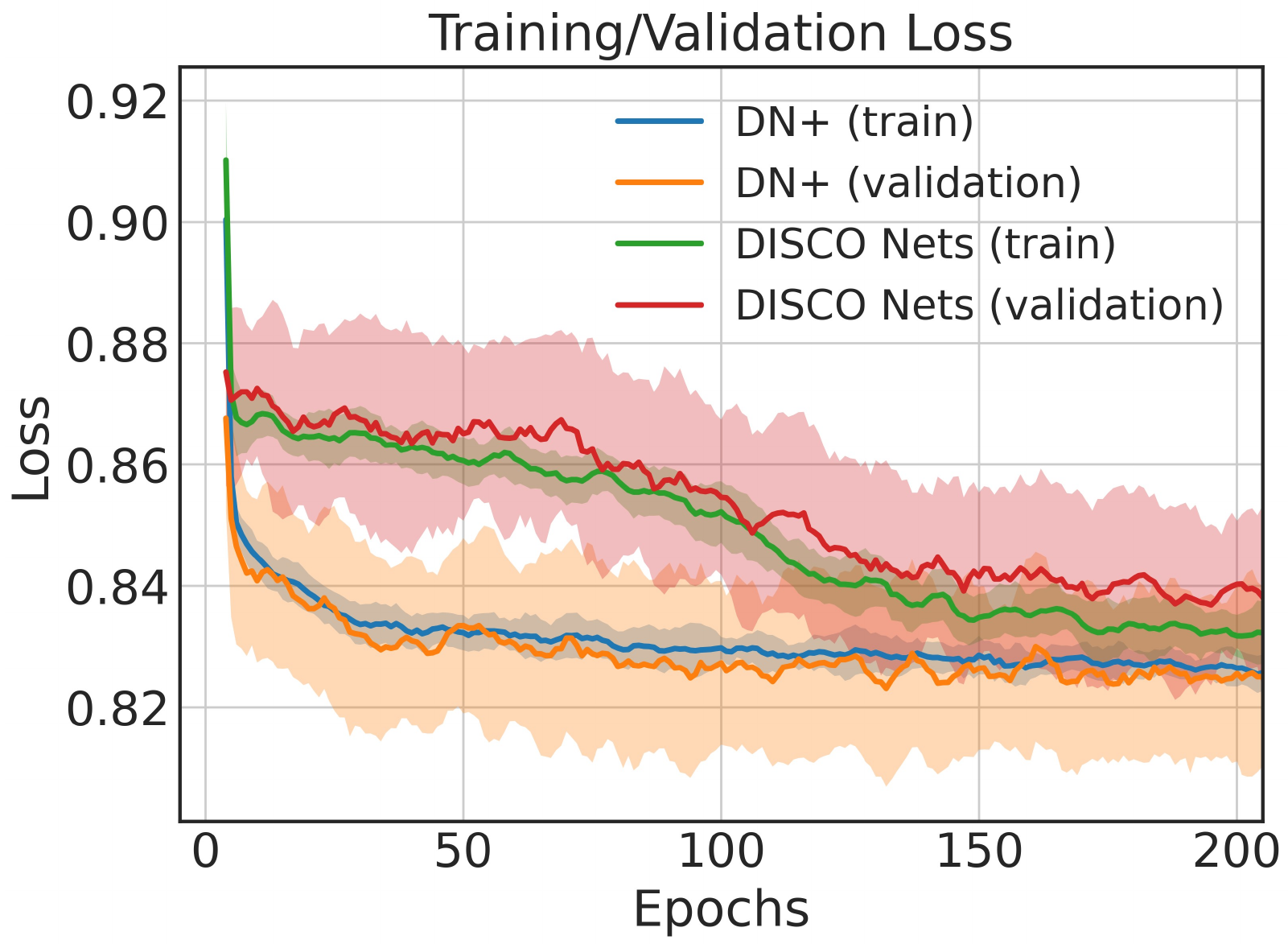}
	\caption{\label{fg:p32rkjdfd}Learning curves of DISCO Nets and {\ALG} trained on the dataset in Figure~\ref{fg:645kojk}. An average over 7 random seeds was taken. The solid lines are mean and the shaded areas are $\pm$ one standard deviation.}
\end{figure}

Finally we make a direct comparison between DISCO Nets \cite{disco2016} and {\ALG}. In the former, the dimension of a random vector $\a$ concatenated with the original input $\x$ is arbitrary and there is no general guiding principle to decide it.%
\footnote{In \cite{disco2016}, DISCO Nets with $\mathrm{dim}\,\a=200$ was used to model a target with $\mathrm{dim}\,\y=42$.} In this comparison we use scalar $a$ in alignment with {\ALG}. We adopt the same hyperparameters (learning rate, batch size, $\Nb$, activation function, hidden layer width) to ensure a fair comparison. We split the dataset into training data (85\%) and validation data (15\%) for the purpose of monitoring the progress of training. The learning curves of both models are displayed in Figure~\ref{fg:p32rkjdfd}. 
We observe that {\ALG} converges substantially faster than DISCO Nets. For instance, {\ALG} takes only 20 epochs to reach the validation loss below $0.84$, whereas DISCO Nets need $150$ epochs, hence {\ALG} achieves more than 7x speedup. We have also experimented with DISCO Nets with a noise vector $\a$ in more than one dimension, finding that the convergence is monotonically delayed as $\mathrm{dim}\,\a$ is increased.

\subsection{Sampling of Functions after Fitting to Data}

\begin{figure}
	\centering
	\includegraphics[width=3.7cm]{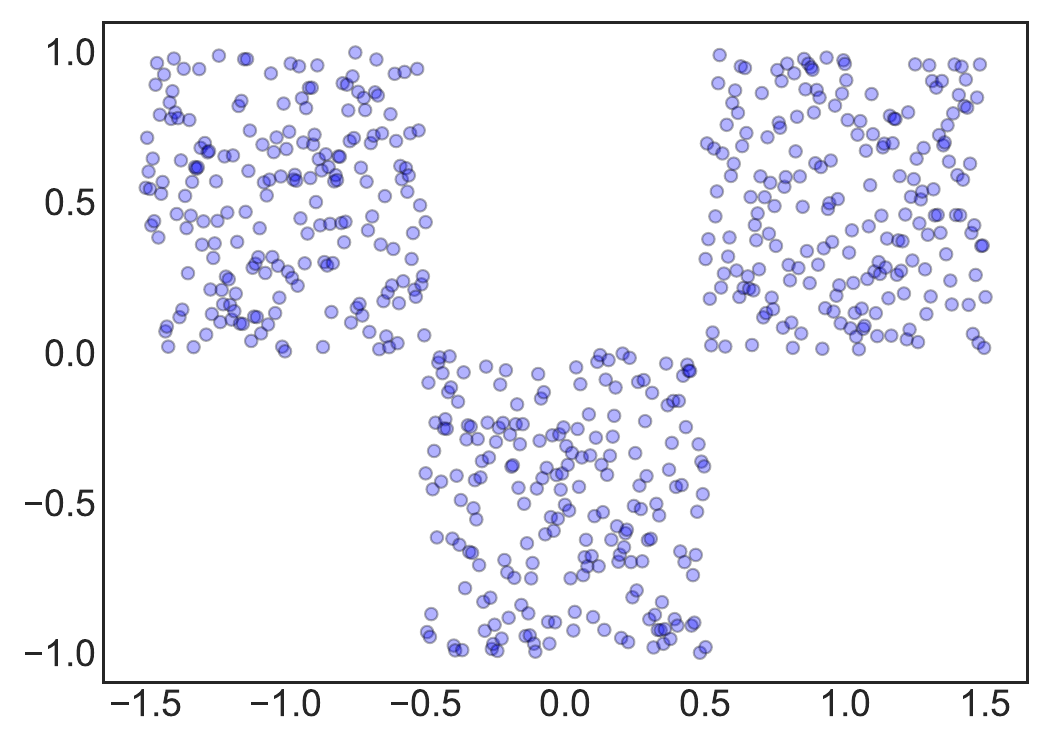}
	\includegraphics[width=.48\textwidth]{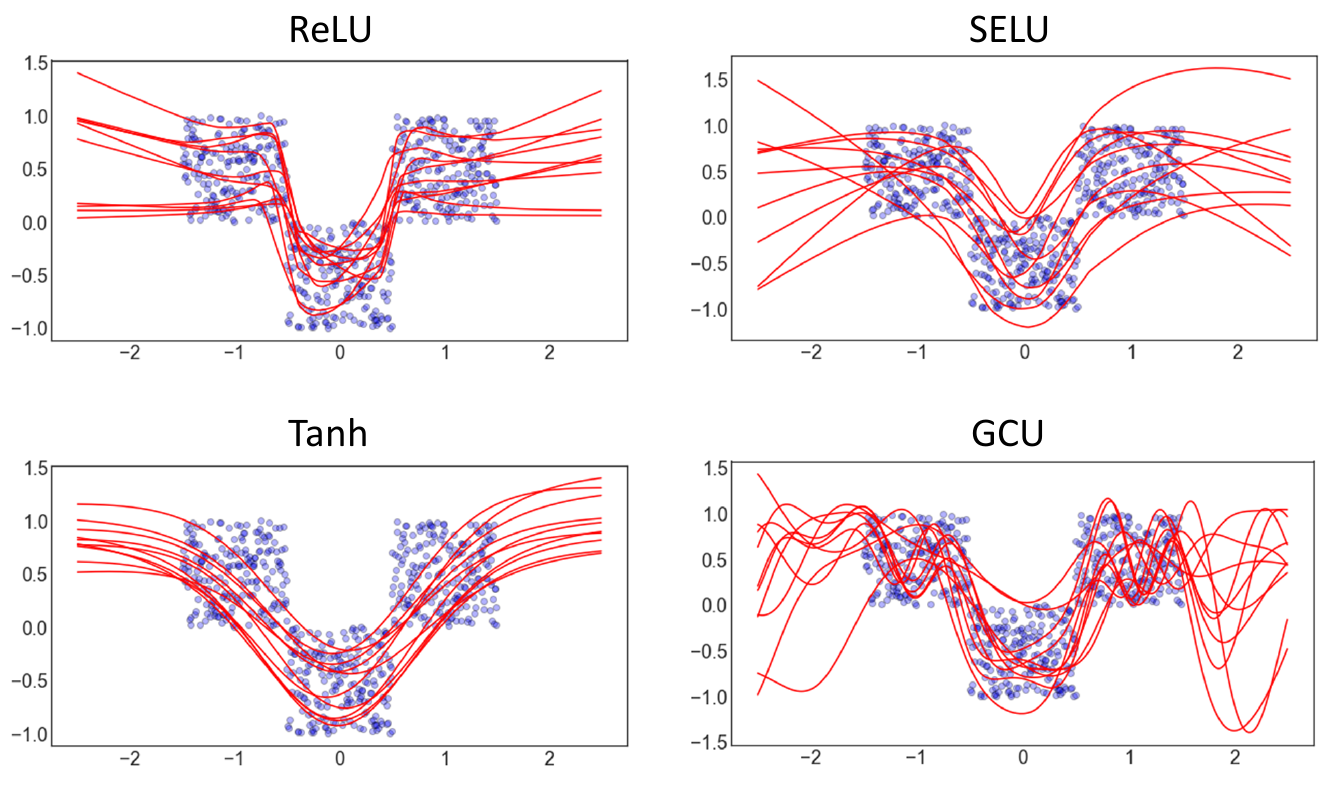}
	\caption{\label{fg:tmnfgoi9}Data distribution (top) and samples drawn from the fitted NN (bottom) for the four activation functions. }
\end{figure}

Drawing samples from a posterior predictive distribution is central to Bayesian inference \cite{Bishop2006}. In GP the cost of naive generation of samples scales cubically with the number of observations, and historically a variety of approximation schemes have been proposed \cite{Lazaro2010,Wilson2020}. In DISCO Nets and {\ALG}, it is in fact trivial to sample functions from the posterior after training on a dataset. While we make an ensemble forecast at given $\x$ by running a forward pass with a large number of distinct inputs $a\sim \Pb$ sampled from the base space $\Xb$, sample functions can be obtained by simply fixing $a\sim \Pb$ and running a forward pass with varying $\x$. For illustration, we generated a test dataset as shown in the top panel of figure~\ref{fg:tmnfgoi9}. In total 600 points were generated uniformly inside three squares of size 1. We have applied {\ALG} to this data with hyperparameters $n_{\rm epoch}=50$, $\Nb=20$ and $b_{\rm batch}=80$ using four activation functions. For each of them we sampled 10 functions, as shown in the bottom panel of figure~\ref{fg:tmnfgoi9}. They cover the range of data in a similar way but perform qualitatively different extrapolations outside the range of data. These characteristics that are specific to activation functions, akin to the choice of kernel functions in GP, must be taken into consideration carefully when using this sampling technique for practical purposes. 

\subsection{Two-dimensional Prediction}

Next, we proceed to considering the case where the response variable $\y$ is two-dimensional. An industrial example is the forecasting of solar and wind power generation. These variables are generally correlated with each other and it is insufficient to make separate predictions for them, highlighting the need for a joint multi-dimensional predictive distribution. 

\begin{figure}
	\centering
	\includegraphics[width=.49\textwidth]{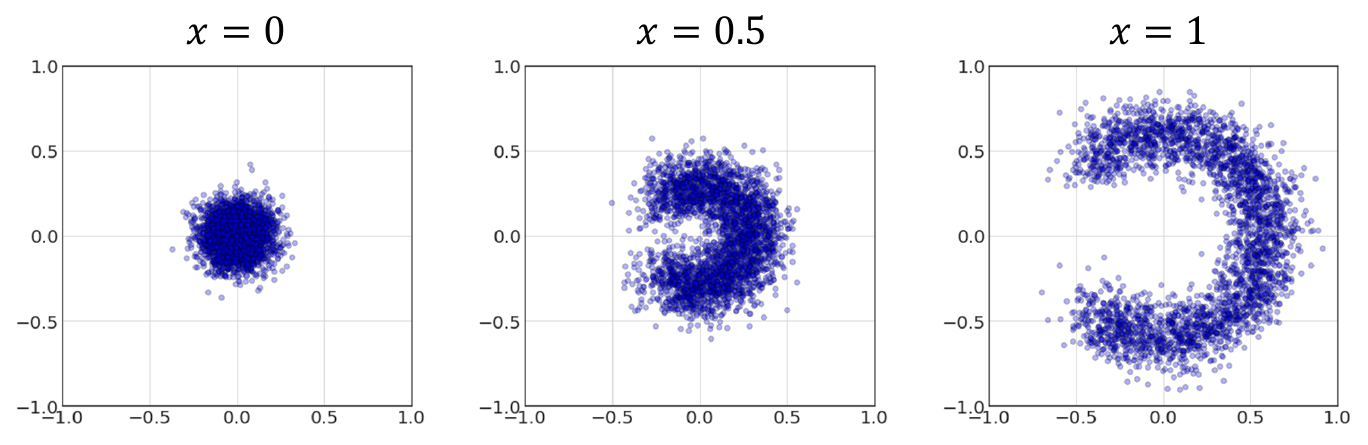}
	\caption{\label{fg:hg564f}The dataset generated for the experiment with 2D output $\y=(y_1,y_2)$.}
\end{figure}

\begin{figure}
	\centering
	\includegraphics[height=3.5cm]{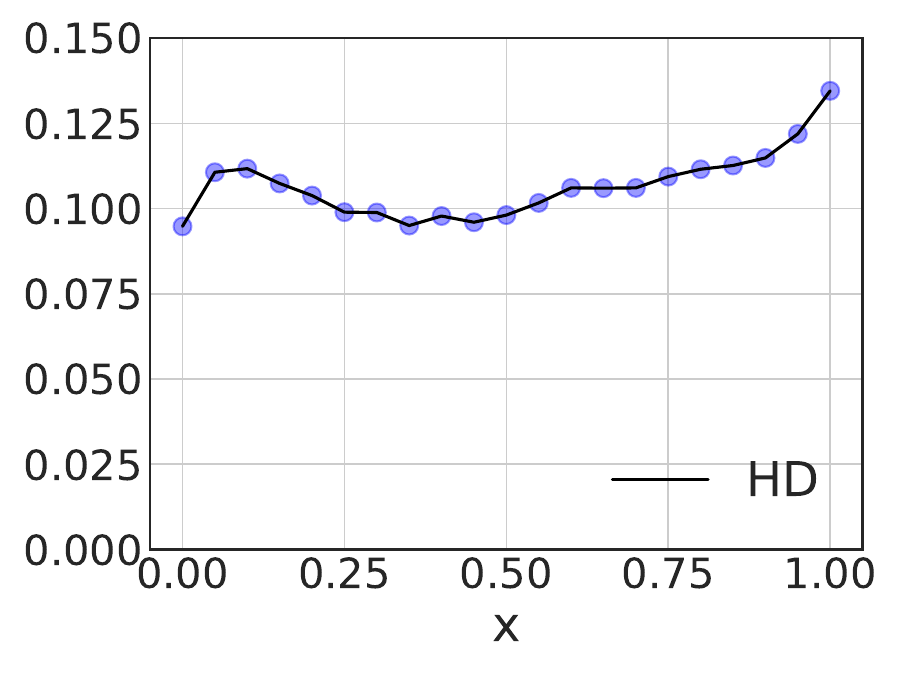}
	\caption{\label{fg:8hfgfd97}Discrepancy between the true distribution $p(\y|x)$ and the predicted distribution $p(\wh{\y}|x)$ measured by the Hellinger distance based on $10^5$ ensemble forecast.}
	\vspace{3mm}
	\centering
	\includegraphics[width=.48\textwidth]{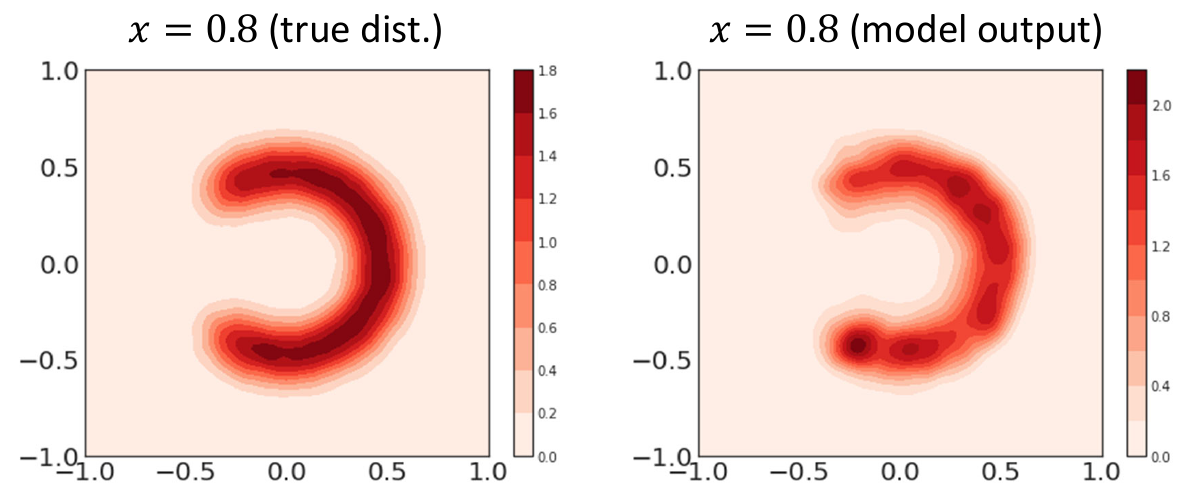}
	\caption{\label{fg:ujhg64}Density plots of the true distribution (left) and {\ALG}'s prediction (right) in the $(y_1,y_2)$-plane.}
\end{figure}

As described in section~\ref{sc:u90gbf}, we sample a two-dimensional random vector $\a\in[0,1]^2$ to make ensemble predictions. As a testbed of the proposed method, we generated a dataset from the distribution below. 
\begin{gather}
	\y=\begin{pmatrix}y_1(x) \\ y_2(x) \end{pmatrix}, \quad 
	\begin{cases}
		y_1(x) = 0.6 \cos \theta \times x + \varepsilon
		\\
		y_2(x) = 0.6 \sin \theta \times x + \varepsilon'
	\end{cases}
	\label{eq:89hgfb912}
	\\
	\varepsilon, \varepsilon' \sim \mathcal{N}\mkakko{0,0.1^2}, \quad 
	\theta \sim U\mkakko{\kkakko{-\frac{3\pi}{4},\frac{3\pi}{4}}} 
\end{gather}
for $0 \leq x \leq 1$. 
The scatter plots of the data at three values of $x$ are displayed in figure~\ref{fg:hg564f}. For small $x$, the data distribution is essentially isotropic, while for larger $x$ a gap on the left gradually opens up. 
The task for {\ALG} is to learn this nontrivial evolution of the distribution over the entire unit interval $x\in [0,1]$. We have taken an equidistant grid of $3000$ points over $[0,1]$ and generated the dataset $\ckakko{(x_i, \y_i)}_{i=1}^{3000}$, which was fed to our NN. We have used ReLU activation and trained the model with $b_{\rm batch}=300$, $\Nb=100$ and $n_{\rm epoch}=300$. 
For various $x$, we made an ensemble forecast with $M=10^5$ and computed its Hellinger distance from the true distribution \eqref{eq:89hgfb912}. The result is shown in figure~\ref{fg:8hfgfd97}. The worst score was $0.135$ at $x=1$ and the mean over all $x$ was $0.106$, hence we conclude that the model has correctly reproduced the data distribution for all $x$. 
For illustration, we show the true distribution and the learned distribution on the $\y$-plane at $x=0.8$ in figure~\ref{fg:ujhg64}. Clearly, major features of the distribution are reproduced to a good accuracy. We anticipate that the result would be improved if the model's layer width is increased further and the hyperparameters of the training such as $b_{\rm batch}$ are optimized.

\subsection{Test on Real-World Datasets\label{sc:2jdtyzdfhb}}

\begin{table*}[ht]
\caption{\label{tb:89bvbcmn}Point forecast performance of each method on each dataset. Best result is in bold face. The lower score is better.}
\centering \hspace*{-7mm}\scalebox{0.8}{
\begin{tabular}{lcccccc|cccccc}
	\toprule
	\multirow{2}{*}{Dataset} 
	& \multicolumn{6}{c}{MAE ($\downarrow$)} 
	& \multicolumn{6}{c}{RMSE ($\downarrow$)}
	\\\cmidrule(lr){2-7}\cmidrule(lr){8-13}
	&QRF&LGBM&NGB&GP&Base NN&
	\colorbox{lightgray}{
		{\ALG}
	}
	&QRF&LGBM&NGB&GP&Base NN&
	\colorbox{lightgray}{
		{\ALG}
	}
	\\\midrule 
	California & 0.345 & 0.328 & {\bfseries 0.319} & 0.368 &\new{0.388}& \old{0.328}\new{0.324}
	& 0.522 & 0.474 & {\bfseries 0.465} & 0.545 &\new{0.614}&\old{0.517}\new{0.501}
	\\
	Concrete &2.63&{\bfseries 2.60}&2.69&2.68&\new{3.05}&\old{2.71}\new{2.73}
	&3.47&{\bfseries 3.35}&3.49&3.61&\new{4.10}&\old{3.65}\new{3.94}
	\\
	Kin8nm&0.116&0.099&0.116&{0.060}&\new{0.061}& \old{0.061}\new{\bfseries 0.057}
	&0.143&0.124&0.150&0.080 &\new{0.081}& \old{{\bfseries 0.079}} \new{\bfseries 0.075}
	\\
	Power Plant &2.37&{\bfseries 2.30} &2.37&2.60&\new{2.82}&\old{2.73}\new{2.67}
	&3.40&{\bfseries 3.22}&3.39&3.63&\new{3.86}&\old{3.76}\new{3.71}
	\\
	House Sales & 68.2k &{\bfseries 63.6k} &68.3k&86.0k&\new{74.9k}&\old{79.7k}\new{71.5k}
	&123k&{\bfseries 111k}&117k&153k&\new{155k}&\old{166k}\new{131k}
	\\
	Elevators &1.87E-3&1.62E-3&1.58E-3&{1.55E-3}&\new{1.51E-3}&\old{1.61E-3}\new{\bfseries 1.43E-3}
	&2.76E-3&2.36E-3&2.15E-3&{2.09E-3}&\new{2.16E-3}&\old{2.24E-3}\new{\bfseries 1.97E-3}
	\\
	Bank8FM &0.0227&{0.0221}&0.0227&0.0231&\new{0.0240}&\old{0.0223}\new{\bfseries 0.0219}
	&0.0311&0.0303&{\bfseries 0.0301}&0.0307&\new{0.0334}&\old{0.0310}\new{0.0305}
	\\
	Sulfur &{\bfseries 0.0172}&0.0194&0.0224&0.0183&\new{0.0204}&\old{0.0197}\new{0.0203}
	&0.0308&0.0348&0.0386&{\bfseries 0.0302}&\new{0.0424}&\old{0.0338}\new{0.0374}
	\\
	Superconduct &{\bf 5.06}&6.45&6.66&5.35&\new{6.05}& \old{5.53} \new{5.46}
	& {\bf 8.96}&10.6&10.5&9.22&\new{10.6}& \old{9.75}\new{10.1}
	\\
	Ailerons &2.07E-4&{\bf 1.07E-4}&2.09E-4&1.15E-4&\new{1.27E-4}&\old{1.30E-4}\new{1.13E-4}
	&2.64E-4&{\bf 1.52E-4}&2.73E-4&1.58E-4&\new{1.80E-4}&\old{1.84E-4}\new{1.58E-4}
	\\
	\midrule 
	\begin{minipage}{15mm}Average\\
	Rank ($\downarrow$)\end{minipage}
	&{3.1}&{\bf 2.6} &{4.0}&{3.5}&\new{4.7}&\old{3.2}\new{2.8}
	&{3.6}&{\bf 2.5} &{3.3}&{3.1}&\new{5.2}&\old{3.5}\new{3.1}
	\\\bottomrule
\end{tabular}
}
\vspace{\baselineskip}\\
\caption{\label{tb:9mleewre}Probabilistic forecast performance of each method on each dataset. Best result is in bold face. The lower score is better.}
\vspace{\baselineskip}
\scalebox{0.85}{
\begin{tabular}{lccccc|ccccc}
	\toprule
	\multirow{2}{*}{Dataset} 
	& \multicolumn{5}{c}{NLL ($\downarrow$)}
	& \multicolumn{5}{c}{CRPS ($\downarrow$)}
	\\\cmidrule(lr){2-6}\cmidrule(lr){7-11}
	&QRF&LGBM&NGB&GP&
	\colorbox{lightgray}{
		{\ALG}
	}
	&QRF&LGBM&NGB&GP&
	\colorbox{lightgray}{
		{\ALG}
	}
	\\\midrule 
	California 
	& 0.316 & 0.194 & 0.111 & 0.411 &\old{\bfseries 0.097}\new{\bfseries 0.077}
	& 0.143 & {\bfseries 0.130} & 0.139 & 0.167 &\old{0.134}\new{0.132}
	\\
	Concrete 
	&2.74&2.44&2.39&2.48&\old{\bfseries 2.36}\new{\bfseries 2.18}
	&1.80&{\bfseries 1.03}&1.39&1.33&\old{1.38}\new{1.09}
	\\
	Kin8nm
	& $-0.590$ & $-0.972$ & $-0.869$ & {$-1.35$} & \old{$-0.964$} \new{$\mathbf{-1.54}$}
	&0.065&0.048&0.058& {0.031} & \old{0.036}\new{\bfseries 0.027}
	\\
	Power Plant 
	&2.27&2.16&{\bfseries 2.11}&2.39&\old{2.33}\new{2.31}
	&1.16&{\bfseries 1.04}&1.14&1.24& \old{1.27}\new{1.25}
	\\
	House Sales 
	&{\bfseries 12.2}&12.3&12.3&12.4&\old{12.4}\new{12.3}
	&28.8k&{\bfseries 26.0k}&28.2k&35.9k& \old{28.2k}\new{26.9k}
	\\
	Elevators 
	& $-4.89$ & $-5.09$ & {$-5.14$} & $-5.04$ & \old{$-5.11$}\new{$\bf -5.23$}
	& 8.76E-4 & {7.56E-4} & 7.80E-4 & 7.65E-4 & \old{8.09E-4}\new{\bf 6.80E-4}
	\\
	Bank8FM 
	& $-2.19$ & $-2.29$ & {$\bf -2.45$} & $-2.39$ & \old{$-2.36$}\new{$-2.31$}
	& 0.0128 & 0.0110 & {\bfseries 0.0105} & 0.0106 & \old{0.0107}\new{0.0110}
	\\
	Sulfur 
	& {$\bf -2.89$} & $-2.79$ & $-2.59$ & $-2.52$ & \old{$-2.72$}\new{$-2.69$}
	& {\bfseries 6.30E-3} & 7.55E-3 & 9.63E-3 &8.90E-3 & \old{7.79E-3}\new{7.94E-3}
	\\
	Superconduct
	&{\bf 2.72}&3.09&3.14&3.32&\old{2.87}\new{2.79}
	&{\bf 1.63}&2.16&2.67&2.76&\old{1.87}\new{1.68}
	\\
	Ailerons 
	& $-7.03$ & $-7.72$ & $-7.06$ & $-7.65$ & \old{$\bf -7.95$}\new{$\bf -7.85$}
	& 9.42E-5 & {\bf 5.17E-5} & 1.21E-4 & 5.55E-5 & \old{6.54E-5}\new{\bf 5.17E-5}
	\\\midrule 
	\begin{minipage}{15mm}Average\\
	Rank ($\downarrow$)\end{minipage} 
	&{3.5}&{2.7}&{2.6}&{4.0}&\old{\bf 2.3}\new{\bf 1.9}
	&{3.7}&{\bf 1.8}&{3.5}&{3.6}&\old{2.9}\new{2.2}
	\\\bottomrule
\end{tabular}
}
\end{table*}

Finally we compare the performance of various baselines with {\ALG} on 10 publicly available tabular datasets for regression tasks. For details of datasets we refer to Appendix~\ref{ap:8uyt5}. This time we have included NGBoost \cite{Duan2020} in the baselines, which fits parametric distributions to data via gradient boosting of decision trees. For each dataset, we have used 90\% of the data for training and 10\% for testing. Furthermore, 10\% of the training data was held out as a validation set, which was used to decide the number of epochs via early stopping. 
Hyperparameters of the baselines are provided in Appendix~\ref{ap:8uyt5}. In {\ALG}, we used the hyperparameters $(b_{\rm batch}, \Nb)=(200,100)$ for all datasets except for the Concrete dataset for which $(b_{\rm batch}, \Nb)=(40,100)$, reflecting the smallness of this dataset. In the inference mode, $M=500$ ensemble predictions were made. {\tt QuantileTransformer(\verb|output_distribution=|\newline 'normal')} from Scikit-Learn was used to preprosess both the input and output of the NN.

As another baseline, we also include Base NN, a plain vanilla NN with no UQ capability. The architecture of Base NN is exactly the same as the left half of Figure~\ref{fg:43rds}; Base NN does not use the external vector $a\sim \Pb$ and simply maps the input $\x_i$ to the output $\y_i$. 

The result on the point forecast performance measured by MAE and RMSE is displayed in table~\ref{tb:89bvbcmn}. Overall, LightGBM was the best performing model while Base NN was the worst. This is partly expected: as is widely recognized, highly tuned gradient boosting models often outperform simple MLP. Next, comparing Base NN with {\ALG}, we notice that {\ALG} has reduced MAE by 8\% and RMSE by 10\% on average. This is a substantial improvement, despite the fact that the architectural changes of {\ALG} over Base NN were purely meant to enable UQ rather than to improve point forecast accuracy. Overall, {\ALG} was second to LightGBM and better than NGBoost, GP, and QRF. As an aside we note that the computational cost for {\ALG} was 12.8 times higher than Base NN on average.

The probabilistic forecast performance is compared in table~\ref{tb:9mleewre}. As a metric we used negative log-likelihood (NLL) and CRPS. In order to compute NLL numerically, we have used Gaussian KDE for the four methods other than NGBoost. This time, {\ALG} was best on average on NLL, while LightGBM was best on average on CRPS, respectively.%
\footnote{In order to generate ensemble forecast, we had to train 51 LightGBM models corresponding to different quantile points, which increased computational cost significantly.} It is noteworthy that {\ALG} was best on five datasets for NLL and three datasets for CRPS. Overall, the probabilistic forecast performance of {\ALG} was at least quite competitive with the standard baselines of probabilistic regression. 

\begin{figure}
	\centering \hspace*{-3mm}
	\includegraphics[width=.5\textwidth]{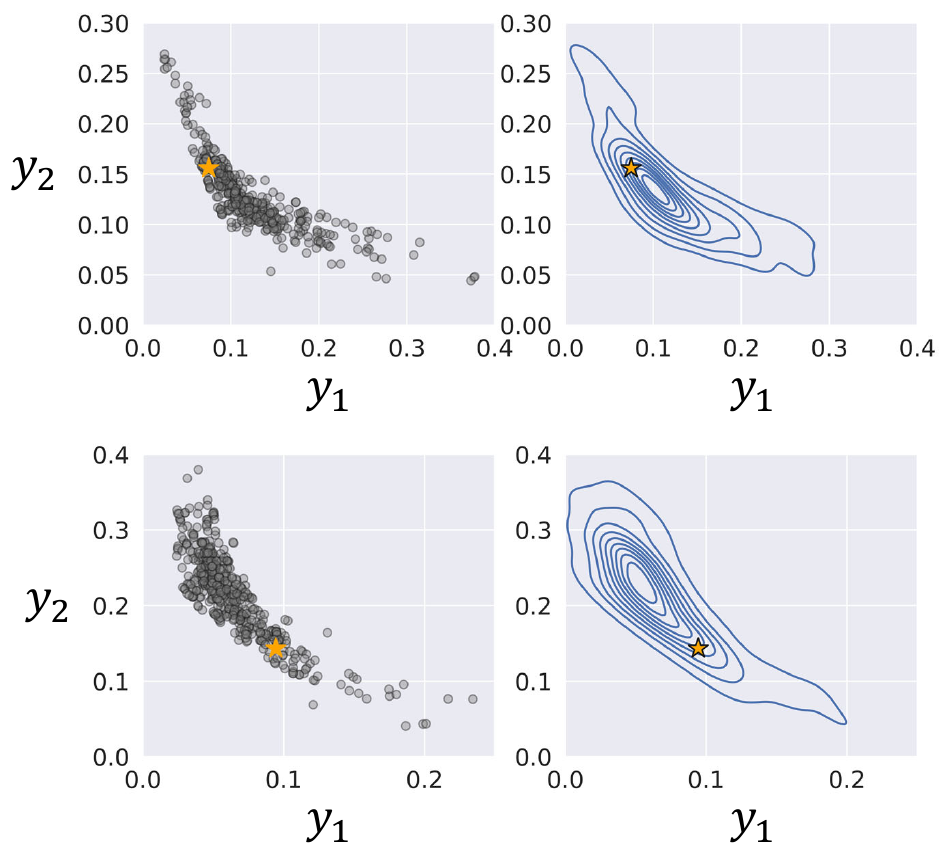}
	\caption{\label{fg:22dd}{\ALG}'s predictive distributions for two samples from Sulfur dataset. The left two panels show 500 ensemble predictions drawn from {\ALG}, from which the contour plots on the right were generated via kernel density estimation, to guide the eye. The orange star in each panel is the ground-truth label of $\y$.}
\end{figure}

When the target variable $\y$ is not a scalar, most of the baselines (QRF, LightGBM, and NGBoost) fail to make a joint probabilistic forecast of multiple components of $\y$, whereas {\ALG} can be applied straightforwardly, as was shown on toy datasets in Figures~\ref{fg:8hfgfd97} and \ref{fg:ujhg64}. To showcase this capability for real-world data, we trained {\ALG} on the Sulfur dataset (cf.~Appendix~\ref{ap:8uyt5}), in which $y_1$ and $y_2$ represent the concentration of \text{H$_2$S} and \text{SO$_2$}, respectively. Two examples of the probabilistic forecast for $(y_1,y_2)$ by {\ALG} are displayed in Figure~\ref{fg:22dd}, where the ground-truth label (a tiny orange star) is overlaied with the density forecast created by sampling 500 predictions from {\ALG}.%
\footnote{Note that the dots in the left panels of Figure~\ref{fg:22dd} are \emph{not} experimental data but the ensemble forecast by {\ALG}.} 
These figures illustrate that {\ALG} is indeed capable of producing a skewed non-Gaussian predictive distribution quite straightforwardly.

\subsection{Explainability}

Explainability and interpretability of a machine-learning model is essential for wider adoption of AI by the society. Diverse methods have been introduced in the research community of eXplainable AI (XAI) \cite{xaibook}. One of the most popular methods to explain an output of a model is to compute feature importance, i.e., the influence of each input feature on the output of a model. So far, the computation of feature importance for distributional forecast has been difficult. Suppose a NN model yields a Gaussian predictive distribution by predicting its mean and variance, $(\mu(\x), \sigma(\x)^2)$. It is then easy to apply any of the existing XAI methods to explain $\mu(\x)$ or $\sigma(\x)$, but this is a rather coarse explanation of the output; recall that the predictive distribution has both a high tail (say, $y>\mu(\x)+2\sigma(\x)$) and a low tail (say, $y<\mu(\x)-2\sigma(\x)$) each of which may well have different explanations in principle. Ideally, we should be able to explain any subregion of the distributional forecast and answer questions like ``Which feature is most responsible for the output provided that $y\in[\mu(\x)-2.3\sigma(\x),\mu-0.4\sigma(\x)]$?'' However, such a flexible XAI method for distributional forecast has not been available.

This problem can be solved by using {\ALG}. Suppose we wish to explain the prediction of {\ALG} given an input $\x$ under the assumption on {\ALG}'s prediction that $\wh\y\in \Omega\subset \RR^d$, where $\Omega$ is any user-specified subset of $\RR^d$ satisfying $\int_\Omega \rmd \wh\y\; p(\wh\y|\x)\ne 0$. First, we randomly draw a finite number (say, 1000) of vectors $\{\a_i\}_i$ from the base distribution $\Pb$, and then compute the ensemble forecast $\{\wh{\y}(\x,\a_i)\}_i$. Next, we form a set $A_\Omega:=\{\a_i \;|\; \wh\y(\x,\a_i)\in\Omega\}$. Finally, we compute the feature importance of NN for each fixed $\a_i\in A_\Omega$ and take their average. To be precise, let $\overrightarrow{\rm FI}(\x,\a_i)$ denote the vector of feature importance computed for a NN that receives $(\x,\a_i)$ as input. (The dimension of $\overrightarrow{\rm FI}$ is the same as that of $\x$.) Then the desired feature importance conditioned on $\Omega$ is simply given by
\ba
	\overrightarrow{\rm FI}(\x | \Omega) = 
	\frac{1}{|A_\Omega|}\sum_{\a_i\in A_\Omega}
	\overrightarrow{\rm FI}(\x,\a_i)\,.
\ea

\begin{figure*}
	\centering 
	\includegraphics[width=\textwidth]{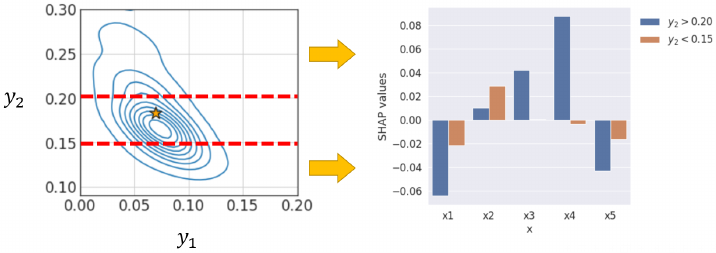}
	\caption{\label{fg:xai}Feature importance (SHAP values \cite{shap2017}) for {\ALG}'s prediction of $y_2$ conditioned on a subregion of the predictive distribution.}
\end{figure*}

As a demonstration of this method, we took a random sample from the Sulfur dataset and computed the feature importance of $\x\in\RR^5$ in two regions specified by $y_2>0.2$ and $y_2<0.15$, respectively, as shown in Figure~\ref{fg:xai}. We used the Kernel SHAP method \cite{shap2017,kernelshap} to compute $\overrightarrow{\rm FI}$. The right panel of Figure~\ref{fg:xai} shows a clear difference between the feature importances in each region. Such a fine-grained explanation of probabilistic forecast will benefit risk-sensitive decision making in real-world applications.

\section{\uppercase{Conclusion}\label{sc:cdosc7897}}

In this paper, we introduced a method {\ALG} built on top of DISCO Net \cite{disco2016} for uncertainty quantification with neural networks for regression tasks. {\ALG} is a nonparametric method that can model arbitrary complex multimodal distributions. In contrast to Bayesian neural networks and Gaussian processes, {\ALG} scales well to a large dataset without a computational bottleneck. It also offers an easy way to sample as many functions as desired from the posterior distribution. Due to its simplicity, {\ALG} can be combined with a variety of complex neural networks, including convolutional and recurrent neural networks. In numerical experiments, we have shown that {\ALG} can even model a two-dimensional distribution of a response variable successfully, in stark contrast to conventional quantile-based approaches that struggle in higher than one dimensions. 

In future work, it would be interesting to apply {\ALG} to uncertainty quantification problems where images or time-series data are given as inputs to the model.

\bibliography{draft_v3.bbl}
\section*{\uppercase{Appendix}}
\setcounter{subsection}{0}
\renewcommand{\thesubsection}{\Alph{subsection}}

\subsection{\label{ap:9hufg}Minimizer of the loss function}
As mentioned in section~\ref{sc:09e224987}, the energy score \eqref{eq:dfsgfd4} is a \emph{strictly proper} scoring rule, making it an ideal tool for distributional learning. However, the underlying mathematics \cite{Szekely2003} may not be easily accessible for practitioners of data science. In this appendix, we present a short and rudimentary argument highlighting that NN training based on the energy score does indeed allow to learn the correct predictive distribution.

Let us begin by observing that, when the loss \eqref{eq:o93cd} is averaged over all data in the minibatch, it approximately holds that 
\ba
	& \quad 
	\frac{1}{b_{\rm batch}}\sum_{k=1}^{b_{\rm batch}}
	\L \mkakko{\y_k, \big\{\wh{\y}_k^{(n)}\big\}_{n=1}^{\Nb}} 
	\notag
	\\
	& \simeq \int \rmd \x \; p(\x) \bigg[
		\int \rmd \y \; p(\y|\x) \int \rmd \wh{\y} \; 
		p(\wh{\y}|\x) \|\y-\wh{\y}\|
	\notag
	\\
	& \quad 
		- \frac{1}{2} \int \rmd \wh{\y}\; p(\wh{\y}|\x)
		\int \rmd \wh{\y}' \; p(\wh{\y}'|\x) \|\wh{\y} - \wh{\y}'\|
	\bigg] 
\ea
where $p(\y|\x)$ is the true conditional distribution and $p(\wh{\y}|\x)$ is the predicted distribution by the model. To stress that they are different functions, we will hereafter write $p(\y|\x)$ as $p_\x(\y)$ and $p(\wh{\y}|\x)$ as $f_\x(\y)$, respectively. Our interest is in the stationary point(s) of the loss function as a functional of $f_\x(\y)$. Before taking the functional derivative w.r.t.~$f_\x$, we must take note of the fact that $\int \rmd \y\; f_\x(\y)=1$. Introducing a Lagrangian multiplier function $\lambda(\x)$, we obtain the loss functional
\ba
	& \int \rmd \x\; p(\x) 
		\int \rmd \y \int \rmd \y' \; 
		\ckakko{p_\x(\y) f_\x(\y')-\frac{1}{2}f_\x(\y)f_\x(\y')}
	\notag
	\\
	& \times \|\y - \y'\| - \int \rmd \x\; 
	\lambda(\x) \mkakko{\int \rmd \y \; f_\x(\y) - 1}.
\ea
Differentiation w.r.t.~$f_\x(\y)$ yields the saddle point equation
\ba
	\int \rmd \y' \big\{p_\x(\y') - f_\x(\y')\big\} 
	\|\y-\y'\| = \frac{\lambda(\x)}{p(\x)} \,.
	\label{eq:9ggd9}
\ea
Let us consider the case where $\y$ is a scalar. Using $\der_y^2 |y-y'|=2\delta(y-y')$ we may take the second derivative of both sides of \eqref{eq:9ggd9} and obtain $f_\x(\y)=p_\x(\y)$. This means that the model's prediction agrees with the true conditional distribution, which is the desired result. 

Actually this mathematical trick applies to arbitrary \emph{odd} dimensions. For illustration, suppose $\y$ is in three dimensions. The radial component of the Laplacian in three dimensions is given by
\ba
	\Delta g(r) = \frac{1}{r^2}\der_r \big(r^2 \der_r g(r)\big)
\ea
for $r=\sqrt{x_1^2+x_2^2+x_3^2}$\;, so
\ba
	\Delta^2 r =\Delta(\Delta r)\propto \Delta \frac{1}{r} \propto \delta(\x). 
\ea
In the last step we have used the fact that $1/r$ is the Green's function of the Poisson equation in three dimensions. Therefore, acting $\Delta_\y$ on both sides of \eqref{eq:9ggd9} yields $f_\x(\y)-p_\x(\y)=0$. 

The case of even dimensions needs a little more caution. Let $\kappa(\a,\b):=\|\a - \b\|$. Suppose $g(\a) = \int \rmd \b\; \kappa(\a,\b)f(\b)$ holds for some functions $f$ and $g$. Then it is easy to verify that $f(\c)=\int \rmd \a\; \kappa^{-1}(\c, \a)g(\a)$, where the inverse of $\kappa$ is defined by the relation 
\ba
	\int \rmd \a\; \kappa^{-1}(\c,\a)\kappa(\a,\b)=\delta(\c-\b)\,.
\ea
Using this for \eqref{eq:9ggd9} we formally obtain
\ba
	p_\x(\y) - f_\x(\y) & = \frac{\lambda(\x)}{p(\x)}
	\int \rmd \y'\; \kappa^{-1}(\y,\y') \,.
	\label{eq:8gfd0d}
\ea
We compute $\kappa^{-1}$ in the momentum space since convolution becomes an algebraic product. Let
\ba
	\|\a \| & = \int \frac{\rmd^d q}{(2\pi)^d}\rme^{i \a \cdot \q}\mu(\q)\,.
\ea
Then
\ba
	\mu(\q) & = \int \rmd \b \; \rme^{-i\q\cdot \b}\|\b\|
	\\
	& = \lim_{\epsilon\to +0} \int \rmd \b \; \rme^{-\epsilon \|\b\| -i\q\cdot \b}\|\b\| \,.
\ea
For $d=2$, a careful calculation shows that
\ba
	\mu(\q) & = - \frac{2\pi}{\|\q\|^3}\,.
\ea
Therefore
\ba
	\kappa^{-1}(\y,\y') & = - \frac{1}{2\pi}
	\int \frac{\rmd^2 p}{(2\pi)^2}\rme^{i(\y-\y')\cdot \p}\|\p\|^3
	\\
	& = \frac{1}{2\pi}\Delta_{\y'}
	\int \frac{\rmd^2 p}{(2\pi)^2}\rme^{i(\y-\y')\cdot \p}\|\p\|
	\\
	& = - \frac{1}{4\pi^2}\Delta_{\y'} 
	\frac{1}{\|\y-\y'\|^3}\,.
\ea
Plugging this into \eqref{eq:8gfd0d} yields $p_\x(\y) = f_\x(\y)$, as desired. Here we have used the fact that the integral of a total derivative vanishes due to the absence of a surface contribution. The argument above generalizes to higher even $d$.

\subsection{\label{ap:8uyt5}Summary of Real-World Datasets }
Below is the summary of the datasets used for the benchmark test in section~\ref{sc:2jdtyzdfhb}. 
\begin{table}[H]
\scalebox{0.75}{
\begin{tabular}{lcccl}
	\toprule 
	Dataset & Size &dim($\x$)& URL
	\\\midrule
	California & 20640 & 8 & 
	\begin{minipage}{.3\textwidth}
	\url{https://scikit-learn.org/stable/modules/generated/sklearn.datasets.fetch_california_housing.html}
	\end{minipage} \\
	Concrete & 1030 & 8 & 
	\begin{minipage}{.3\textwidth}
	\url{https://www.openml.org/d/4353}
	\end{minipage}\\
	Kin8nm &8192&8 & \begin{minipage}{.3\textwidth}
	\url{https://www.openml.org/d/189}
	\end{minipage}\\
	Power Plant &9568& 4 & \begin{minipage}{.3\textwidth}
	\url{https://archive.ics.uci.edu/ml/datasets/combined+cycle+power+plant}
	\end{minipage}\\
	House Sales & 21613 & 19 & \begin{minipage}{.3\textwidth}
	\url{https://www.openml.org/d/42635}
	\end{minipage}\\
	Elevators & 16599 & 18 & \begin{minipage}{.3\textwidth}
	\url{https://www.openml.org/d/216}
	\end{minipage}\\
	Bank8FM & 8192 & 8 & \begin{minipage}{.3\textwidth}
	\url{https://www.openml.org/d/572}
	\end{minipage}\\
	Sulfur & 10081 & 5 & \begin{minipage}{.3\textwidth}
	\url{https://www.openml.org/d/23515}
	\end{minipage}\\
	Superconduct & 21263 & 81 & \begin{minipage}{.3\textwidth}
	\url{https://www.openml.org/d/43174}
	\end{minipage}\\
	Ailerons & 13750 & 33 & \begin{minipage}{.3\textwidth}
	\url{https://www.openml.org/d/44137}
	\end{minipage}\\\bottomrule
\end{tabular}
}
\end{table}
Remarks:
\begin{itemize}
	\item The number of trees in QRF was set to 1000. 
	\item The number of trees in LightGBM was set to 300. {\tt \verb|learning_rate|} and {\tt \verb|min_child_weight|} were tuned.
	\item The number of trees in NGBoost was set to 500. {\tt \verb|learning_rate|} was tuned. The Gaussian distribution was used for fitting.
	\item GP's kernel used the sum of the isotropic Matern 
\mbox{kernel and the white kernel.~{\tt  QuantileTransformer}}
	\newline{\tt (\verb|output_distribution|='normal')} from Scikit-Learn was used to preprosess the input, and {\tt \verb|RobustScaler()|} from Scikit-Learn was used to scale the output.
	\item To calculate CRPS we have used the library \texttt{properscoring} available at \url{https://github.com/TheClimateCorporation/properscoring}.
\end{itemize}

\end{document}